\documentclass[11pt]{article}
\usepackage{array}
\newcolumntype{P}[1]{>{\centering\arraybackslash}p{#1}}
\usepackage[a4paper, total={6in, 9in}]{geometry}
\usepackage[T1]{fontenc}
\usepackage{graphicx}
\usepackage{multirow}
\usepackage{booktabs}
\usepackage{amsmath}
\usepackage{booktabs,caption}
\usepackage[flushleft]{threeparttable}
\usepackage[displaymath]{lineno}
\usepackage{setspace}
\usepackage{comment}
\usepackage{hyperref}
\usepackage{cleveref}
\usepackage{placeins}
\usepackage[font=small,labelfont=bf]{caption} 
\usepackage[subrefformat=parens]{subcaption}
\usepackage{soul}
\usepackage{color}
\usepackage{amsthm}
\usepackage{mathtools}
\usepackage{pdflscape}
\usepackage{rotating}
\usepackage{mathtools,bbm}
\usepackage{longtable}
\usepackage{float}

\begin{document}

\title{\textbf{Comparing hundreds of machine learning and discrete choice models for travel demand modeling: an empirical benchmark}}

\author{
   Shenhao Wang\textsuperscript{1,4,5}\footnote{Corresponding Author; Email: shenhao@mit.edu}, Baichuan Mo\textsuperscript{3}, Yunhan Zheng\textsuperscript{3, 5}, Stephane Hess\textsuperscript{2}, Jinhua Zhao\textsuperscript{5} \\
  \\
  \small{1 - Department of Urban and Regional Planning, University of Florida} \\
  \small{2 - Choice Modeling Centre \& Institute for Transport Studies, University of Leeds} \\
  \small{3 - Department of Civil and Environment Engineering, Massachusetts Institute of Technology} \\
  \small{4 - Media Lab, Massachusetts Institute of Technology} \\
  \small{5 - Department of Urban Studies and Planning, Massachusetts Institute of Technology} 
}
\date{}              

\renewcommand{\thefootnote}{\fnsymbol{footnote}}
\singlespacing
\maketitle

\vspace{-.2in}
\begin{abstract}
\noindent
Numerous studies have compared machine learning (ML) and discrete choice models (DCMs) in predicting travel demand. However, these studies often lack generalizability as they compare models deterministically without considering contextual variations. To address this limitation, our study develops an empirical benchmark by designing a tournament model to learn the intrinsic predictive values of ML and DCMs. This novel approach enables us to efficiently summarize a large number of experiments, quantify the randomness in model comparisons, and use formal statistical tests to differentiate between the model and contextual effects. This benchmark study compares two large-scale data sources: a database compiled from literature review summarizing 136 experiments from 35 studies, and our own experiment data, encompassing a total of 6,970 experiments from 105 models and 12 model families, tested repeatedly on three datasets, sample sizes, and choice categories. This benchmark study yields two key findings. Firstly, many ML models, particularly the ensemble methods and deep learning, statistically outperform the DCM family and its individual variants (i.e., multinomial, nested, and mixed logit), thus corroborating with the previous research. However, this study also highlights the crucial role of the contextual factors (i.e., data sources, inputs and choice categories), which can explain models' predictive performance more effectively than the differences in model types alone. Model performance varies significantly with data sources, improving with larger sample sizes and lower dimensional alternative sets. After controlling all the model and contextual factors, significant randomness still remains, implying inherent uncertainty in such model comparisons. Overall, we suggest that future researchers shift more focus from context-specific and deterministic model comparisons towards examining model transferability across contexts and characterizing the inherent uncertainty in ML, thus creating more robust and generalizable next-generation travel demand models. 

\hfill\break%
\end{abstract}

\medskip

\thispagestyle{empty}
\clearpage

\onehalfspacing
\setcounter{footnote}{0}
\renewcommand{\thefootnote}{\arabic{footnote}}
\setcounter{page}{1}

\section{Introduction}
\label{sec:introduction}
\noindent
In recent years, there has been a growing interest in incorporating machine learning (ML) models into travel demand analysis, alongside the traditional choice models (DCMs) that have been widely used for decades \cite{Ben_Akiva1996, Cantarella2005, Small2007td, Ortuzar2011}. These studies seek to compare the differences in model performance, system robustness, and ease of interpretation between ML and DCMs \cite{Hagenauer2017, WangFangru2018, ChengLong2019}. These comparative studies play a crucial role in generating valuable insights into novel behavioral mechanisms and diversifying travel demand models in practical applications \cite{WangShenhao2020_econ_info, WangShenhao2020_asu_dnn}. 

However, existing comparative studies suffer from limitations that hinder their robustness and generalizability. They often draw definitive conclusions about the superiority of ML models over DCMs based on a limited number of models and datasets that are specific to particular contexts. Such an approach overlooks the significant sources of randomness in model performance, resulting in potentially misleadingly high confidence in the superiority of ML models. What further adds to the limitations is an implicit assumption that model choice alone drives performance, while disregarding the impact of random contextual factors such as sample sizes and input/output dimensions. To address these shortcomings, it is essential to strive for comparative studies that provide generalizable insights and quantify the confidence in performance comparisons between ML models and DCMs. Unfortunately, past studies have not yet achieved these goals, underscoring the critical need for a benchmark research to overcome these limitations and provide a comprehensive and robust evaluation of the comparative performance of ML models and DCMs in travel demand modeling. 

To address the limitations, we develop a large-scale empirical benchmark by designing a tournament model to compare the performance of hundreds of ML and DCMs.
The tournament model uses the winning model in an experiment as the output, and alternative models and contextual factors as the inputs\footnote{\emph{Inputs} and \emph{outputs} are commonly used in the ML literature, and they correspond to \emph{independent variables} and \emph{dependent variables} in choice modeling.}. It captures the essence of comparative experiments by conducting pairwise comparison for the intrinsic predictive values of models using statistical methods. With a formal statistical approach, our benchmark study can efficiently summarize findings, differentiate between the model and contextual effects, and highlight the shared patterns from a large number of comparative experiments. With the tournament model, this benchmark study can filter out the context-specific noises and reveal common patterns in a specific topic, resembling a conventional meta-analysis while strengthening it with our own large-scale experiments \cite{Ewing2010, Holmgren2007, Patricia2009}. As far as we know, such an empirical benchmark is still missing for travel demand modeling.

This benchmark study leverages two large-scale data sources: data compiled from literature review and from our own experiments. We documented choice categories, sample sizes, alternative models, best models, and many other contextual factors from 136 experiments in 35 past studies, constituting the literature data. This literature data facilitates the design of our 6,970 experiments and the analysis of the large-scale experiment data, in which we compared 105 models from 12 model families, repeatedly on 3 data sets, 3 sample sizes, and 3 outputs. 
The two data sets are collected, processed, and analyzed separately, but their results from the tournament models are compared in the result section. The literature and experiment data are juxtaposed because of their complementarity. The experiments prevail in the completeness of models, while the literature data is rich in contextual diversity. By comparing the literature and experiment data, this benchmark research can identify the consistent and inconsistent patterns. The patterns consistent between literature and experiment data could have more validity than the findings from only literature or experiment data, thus enhancing the generalizability of our empirical findings. The patterns inconsistent between literature and experiment data could reveal potential publication biases, arising from past authors' or journals' preferences for specific findings.


This paper is organized as follows. Section \ref{sec:meta_anlaysis} introduces the design of the tournament model. Section \ref{sec:lr} and Section \ref{sec:experiment_design} introduce the literature and experiment data, and discusses the model and contextual factors. Section \ref{sec:results} presents the empirical findings, starting with the results from the tournament models and further delving into each contributing factor to model performance. Section \ref{sec:limitations} presents limitations, and Section \ref{sec:conclusions} concludes our findings and discuss future research directions. The acronyms are summarized in Appendix I to facilitate the reading of this study. 

\section{Empirical benchmark with a tournament model}
\label{sec:meta_anlaysis}
\noindent The tournament model compares $M$ models, where model $m \in \{1, 2, ..., M\}$ has an intrinsic predictive value $\Tilde{x}^*_m$. This intrinsic predictive value can be decomposed as $\Tilde{x}^*_m = x^*_m + \epsilon_m $, in which $x^*_m$ represents the deterministic component and $\epsilon_m$ the random component. The tournament model compares a pair of such models $m_0 $ and $m_1$ with $m_0, m_1 \in \{1, 2, ..., M\}$. If the intrinsic predictive value of model $m_0$ is higher than model $m_1$, then it is more likely for model $m_0$ to win the comparison with the following probability:
\begin{equation}
    P(m_0 \succ m_1) = \sigma(x^*_{m_0}, x^*_{m_1}) 
\end{equation}
in which $x^*_{m_0}$ and $x^*_{m_1}$ are the deterministic intrinsic predictive values of models $m_0$ and $m_1$, and $\sigma$ takes the form of a link function. Under the assumption of extreme value distributions for $\epsilon_m$, the link function takes the logistic form: 
\begin{align}
    \sigma(x^*_{m_0}, x^*_{m_1}) = \frac{e^{x^*_{m_0}}}{e^{x^*_{m_0}} + e^{x^*_{m_1}}},
\end{align}
guaranteeing that $\sigma(x^*_{m_0}, x^*_{m_1}) + \sigma(x^*_{m_1}, x^*_{m_0}) = 1$. Using $i$ to indicate the comparison between model pairs, the output of each comparison is $y_i \in \{0,1\}$ with $0$ representing the winning of model $m_0$ and $1$ representing the winning of model $m_1$, or vice versa. The winner of each model pair is determined by their prediction accuracy (i.e., hit rates), which allows broad comparison across deterministic (e.g. SVM) and probabilistic (e.g., neural networks) models. After observing a large number of such comparisons, we could infer $x^*_m$ using statistical methods.

It is highly likely that the intrinsic predictive values of model $m$ varies with contextual factors, such as sample size and choice categories. To incorporate the contextual factors, $x^*_m$ can be parameterized as:
\begin{equation}
    x_{m_i}^* = \sum_{\Tilde{\mathcal{M}}\in\mathcal{D}} \beta_{\Tilde{\mathcal{M}}} \cdot \mathbbm{1}_{{m_i} \in \Tilde{\mathcal{M}}}  + \boldsymbol{\gamma}' \boldsymbol{z}_{m_i}
\label{eq:intrinsic_prediction_linear}
\end{equation}
in which $\mathbbm{1}_{{m_i} \in \Tilde{\mathcal{M}}}$ represents the indicators for whether the model $m_i$ belongs to model family $\Tilde{\mathcal{M}}$ with an associated parameter $\beta_{\Tilde{\mathcal{M}}}$, and $\boldsymbol{z}_{m_i}$ represents the contextual factors of model $m_i$ with an associated vector of parameters $\gamma$. The set of all model families is $\mathcal{D}$. We distinguish between model $m_i$ and model family $\Tilde{\mathcal{M}}$; for example, a deep neural network with a specific architecture is a model, while DNNs with various architectures belong to the same model family. In our experiments, we analyze 105 models ($M$ = 105) and 12 model families ($|\mathcal{D}| = 12$). Since the number of model families is significantly smaller, it is more efficient to estimate the coefficients for model families rather than for individual models. 

Equation \ref{eq:intrinsic_prediction_linear} can be further enriched by specifying the interaction and nonlinear contextual effects, such as 
\begin{equation}
    x_{m_i}^* = \sum_{\Tilde{\mathcal{M}}\in\mathcal{D}} \beta_{\Tilde{\mathcal{M}}} \cdot \mathbbm{1}_{{m_i} \in \Tilde{\mathcal{M}}} 
    + \boldsymbol{\gamma'}_{1,\Tilde{\mathcal{M}}} (z_{m_i} \cdot \mathbbm{1}_{{m_i} \in \Tilde{\mathcal{M}}}) 
    + \boldsymbol{\gamma'}_{2,\Tilde{\mathcal{M}}} (z^2_{m_i} \cdot \mathbbm{1}_{{m_i} \in \Tilde{\mathcal{M}}})
\label{eq:intrinsic_prediction_nonlinear}
\end{equation}
in which $\boldsymbol{\gamma}_{1,\Tilde{\mathcal{M}}}$ and $\boldsymbol{\gamma}_{2,\Tilde{\mathcal{M}}}$ capture the linear and quadratic contextual effects specific to model families. This equation is more general than Equation \ref{eq:intrinsic_prediction_linear} because it releases the constraint of equal contextual effects across model families. To fully identify the parameters, our tournament models use discrete choice models (DCMs) as the baseline, so $\beta$'s and $\gamma$'s are interpreted as measuring the differences of intrinsic predictive values between ML models and DCMs.

This tournament model resembles a choice model with alternative sets varying across choice situations. The intrinsic predictive value could be seen as the ``utility'' of model $m$, and in every game (i.e., comparison), a better model $m$ is chosen from a pair of alternatives $m_0$ and $m_1$, which are randomly sampled from a large model set $\{0, 1, ..., M\}$. Therefore, maximum likelihood estimation (MLE) could be adopted to estimate the parameters in the tournament model. 
\begin{equation}
    LL(g(\mathbbm{1}_{{m_i} \in \Tilde{\mathcal{M}}}, z; \beta, \gamma)) = \sum_{(m_j,m_k) \subset \{1, 2, ... M\} } W_{m_j \succ m_k} \log[\sigma(x^*_{m_j}, x^*_{m_k})]
\end{equation}
in which $W_{m_j \succ m_k}$ represents the counts of model $m_j$ outperforming model $m_k$ in the re-sampled model pairs. Hence in the tournament model, the unit of analysis is a re-sampled model pair, rather than a single model. Within a model pair, the two models are compared by their prediction accuracy. Notably, the model pairs are sampled from either the literature or the experiment data, without any cross-comparison between the two data groups.



The following sections comprise two steps for utilizing the tournament model. Firstly, we gather information from individual experiments in the literature, documenting model performance (i.e., prediction accuracy) and corresponding model-family and contextual factors. Simultaneously, we conduct our own experiments, structured similarly to the literature data. In the second step, we randomly select pairs of models separately for the experiment and literature data to compare performance and record the winning model. The pairwise comparisons are then used in the tournament models for subsequent statistical analysis.

\section{Literature data}
\label{sec:lr}
\noindent We compile the literature data by organizing the past studies along several dimensions as shown in Table \ref{table:list_studies}. The dimensions are categorized into the treatment and control dimensions, corresponding to our analytical process in the result section. The treatment dimension includes the compared models (Column 5) and the best model (Column 6), and the control dimensions include the choice categories (Column 3), sample sizes (Column 4), among many other contextual factors, which are not presented in Table \ref{table:list_studies}. The collection process and description of the literature data are incorporated in Appendix IV. 

\footnotesize
\begin{longtable}{p{0.08\linewidth} p{0.28\linewidth} p{0.08\linewidth} p{0.08\linewidth} p{0.25\linewidth} p{0.08\linewidth}}
        \toprule
        \textbf{1. Study Index} & \textbf{2. Author (Year)} & \textbf{3. Choice Types} & \textbf{4. Sample Size} & \textbf{5. Number of Experiments - Compared Models} & \textbf{6. Best Model} \\
        \midrule
        1 & Nijkamp et al. (1996) \cite{Nijkamp1996} & MC & $1,396$ & 2 - DNN, MNL & DNN \\
        2 & Rao et al. (1998) \cite{Rao1998} & MC & $4,335$ & 2 - DNN, MNL & DNN \\
        3 & Hensher and Ton (2000) \cite{Hensher2000} & MC & $1,500$ & 2 - DNN, NL & DNN/NL \\
        4 & Sayed and Razavi (2000) \cite{Sayed2000} & MC & $7,500$ & 2 - MNL, DNN & DNN \\
        5 & Cantarella et al. (2002) \cite{Cantarella2002} & MC & $2,808$ & 3 - DNN, MNL, NL & DNN \\
        6 & Mohammadian et al. (2002) \cite{Mohammadian2002} & CO & $597$ & 2 - DNN, NL & DNN \\
        7 & Doherty and Mohammadian (2003) \cite{Doherty2003} & TP & $5,583$ & 2 - DNN, GLM & DNN \\
        8 & Xie et al. (2003) \cite{XieChi2003} & MC & $4,747$ & 3 - DT, DNN, MNL & DNN \\
        9 & Cantarella et al. (2005) \cite{Cantarella2005} & MC & $1,067$ & 2 - DNN, MNL & DNN \\
        10 & Celikoglu (2006) \cite{Celikoglu2006} & MC & $381$ & 4 - DNN, MNL & DNN \\
        11 & Tortum et al. \cite{Tortum2009} & MC & $441$ & 3 - MNL, GLM, DNN & DNN \\
        12 & Zhang and Xie (2008) \cite{ZhangYunlong2008} & MC & $1,000$ & 3 - MNL, DNN, SVM & SVM \\
        12 & Zhang and Xie (2008) \cite{ZhangYunlong2008} & MC & $2,000$ & 3 - MNL, DNN, SVM & SVM \\
        13 & Biagioni et al. (2008) \cite{Biagioni2008} & MC & $19,118$ & 4 - MNL, BM, BOOSTING, DT & BOOSTING \\
        14 & Xian and Jian (2011) \cite{XianYu2011} & MC & $4,725$ & 3 - SVM, NL, DNN & SVM \\
        15 & Allahviranloo and Recker (2013) \cite{Allahviranloo2013} & TP & $3,671$ & 2 - SVM, MNL & SVM \\
        16 & Omrani et al. (2013) \cite{Omrani2013} & MC & $3,673$ & 6 - DNN, SVM, KNN, MNL, DT, BM & DNN \\
        17 & Ermagun et al. (2015) \cite{Ermagun2015} & MC & $4,700$ & 2 - NL, RF & RF \\
        18 & Jahangiri et al. (2015) \cite{Jahangiri2015} & MC & N.A. & 6 - KNN, SVM, DT, BAGGING, RF, MNL & RF \\
        19 & Tang et al. (2015) \cite{Tang2015} & MC & $72,536$ & 2 - DT, MNL & DT \\
        20 & Omrani (2015) \cite{Omrani2015} & MC & $9,500$ & 4 - DNN, RBFNN, MNL, SVM & DNN \\
        21 & Shafique et al. (2015) \cite{Shafique2015} & MC & $1,968$ & 4 - SVM, BOOSTING, DT, RF & RF \\
        21 & Shafique et al. (2015) \cite{Shafique2015} & MC & $1,488$ & 4 - SVM, BOOSTING, DT, RF & RF \\
        21 & Shafique et al. (2015) \cite{Shafique2015} & MC & $2,754$ & 4 - SVM, BOOSTING, DT, RF & RF \\
        22 & Shukla et al. (2015) \cite{Shukla2015} & MC & $100,000$ & 2 - DNN, DT & DNN \\
        23 & Sekhar and Madhu (2016) \cite{Sekhar2016} & MC & $4,976$ & 3 - RF, DT, MNL & RF \\
        24 & Hagenauer and Helbich (2017) \cite{Hagenauer2017} & MC & $230,608$ & 8 - MNL, DNN, NB, SVM, CTs, BOOSTING, BAGGING, RF & RF \\
        25 & Paredes et al. (2017) \cite{Paredes2017} & CO & $15,211$ & 5 - MNL, BOOSTING, DT, SVM, RF & RF \\
        26 & Hillel et al. (2018) \cite{Hillel2018} & MC & N.A. & 8 - DNN, BAGGING, BOOSTING, KNN, GLM, BM, RF, SVM & BOOSTING \\        
        27 & Golshani et al. (2018) \cite{Golshani2018} & MC & $9,450$ & 2 - MNL, DNN & DNN \\        
        28 & Tang et al. (2018) & MC & $14,000$ & 2 -MNL, DT & DT \\
        29 & Wang and Ross (2018) \cite{WangFangru2018} & MC & $51,910$ & 2 - BOOSTING, MNL & BOOSTING \\
        30 & Lee et al. (2018) & MC & $4,764$ & 2 - MNL, DNN & DNN \\
        31 & Cheng et al. (2019) \cite{ChengLong2019} & MC & $7,276$ & 4 - RF, SVM, BOOSTING, MNL & RF \\
        32 & Zhou et al. (2019) \cite{ZhouXiaolu2019} & MC & $30,000$ & 8 - MNL, KNN, DT, SVM, BM, BOOSTING, BAGGING, RF & BAGGING \\ 
        33 & Wang et al. (2020) \cite{WangShenhao2020_asu_dnn} & MC & $8,418$ & 8 - MNL, NL, DNN, SVM, BM, KNN, BOOSTING, GLM & DNN \\ 
        33 & Wang et al. (2020) \cite{WangShenhao2020_asu_dnn} & MC & $2,929$ & 8 - MNL, NL, DNN, SVM, BM, KNN, BOOSTING, GLM & DNN \\ 
        34 & Wang et al. (2020) \cite{WangShenhao2020_econ_info} & MC & $80,000$ & 2 - MNL, DNN & DNN \\ 
        34 & Wang et al. (2020) \cite{WangShenhao2020_econ_info} & MC & $8,418$ & 2 - MNL, DNN & DNN \\ 
        35 & Wang et al. (2020) \cite{WangShenhao2020_rpsp} & MC & $8,418$ & 3 - NL, MNL, DNN & DNN \\ 
        \midrule 
        \multicolumn{6}{l}{Notes: MC - mode choice; CO - car ownership; TP - trip purposes. Each row is not a study, but an experiment that } \\ 
        \multicolumn{6}{l}{compares models conditioning on the same choice category and sample size} \\ 
        \bottomrule 
\caption{Literature data of 35 studies and 136 experiments}
\label{table:list_studies} 
\end{longtable} 
\normalsize 

Table \ref{table:list_studies} can inform the design of our own experiments to complement the existing studies. In the past literature, the common models include MNL, NL, DNN, GLM, BM, BOOSTING, BAGGING, DT, RF, and SVM. The contextual factors include data sets, sample sizes, and choice categories. The sample size ranges from 381 to 230,608 observations\footnote{The observations are the trips, so the number of individuals and households is smaller.}, and choice categories include travel mode choice, car ownership, and trip purposes. These values are used as baselines for our own experiment design. However, the literature table also demonstrates the weaknesses of past studies. Each study incorporated only a small number of models, and typically failed to recognize the richness within the model families. For example, DNNs can take a variety of architectures and hyperparameters, so a single DNN architecture may not represent the performance of a vast DNN model family. Past studies also miss some important models. On the ML side, the missing models are discriminant analysis (DA) and generalized linear models (GLM). On the DCM side, the missing models are the mixed logit model (MXL), which is a workhorse in practical travel demand modeling. Most importantly, such context-specific studies cannot reveal the general trends and the variations across these findings, which shall be addressed through this benchmark study.

\section{Experimental design}
\label{sec:experiment_design}
\noindent Informed by the literature data, our experiments expand upon four major dimensions ($|H| = 4$), including 105 models ($|T_m|=105$) belong to 12 model families, 3 data sets ($|T_s|=3$), 3 sample sizes ($|T_n|=3$), and 3 choice categories ($|T_y|=3$). In total, it leads to the number of experiment points given by $T = \prod_{h \in H} |T_h|; \ h \in H = \{m,s,n,y\}$, in which $m$ represents models; $s$ datasets; $n$ sample sizes; $y$ outputs; $T$ the total number of experiments; and $T_h$ the cardinality of each dimension. Roughly speaking, our experiments can be treated as a grid search along the four dimensions. Among the four dimensions, the models are the main factor of interest, because the majority of the past studies focused on model comparison. The other three dimensions are contextual factors. 

\subsection{Treatment dimension: models and model families}
\noindent
Table \ref{table:list_classifiers} summarizes the list of 105 individual models (Column 1) from 12 model families (Column 2). Instead of analyzing single models, we create a two-tier structure for models and model families, thus recognizing the richness within each model family. For example, the DNN family includes 16 DNNs with different architectures, thus enabling us to analyze the variation across the DNN architectures. While far from exhaustive, Table \ref{table:list_classifiers} has included the relatively representative models and the most relevant ones for travel behavioral analysis. For example, DCMs are highly relevant given their consistent use in the travel demand analysis, and DNNs are also highly relevant due to their rising popularity in many subdomains in transportation \cite{Karlaftis2011, WangShenhao2020_asu_dnn, WangShenhao2020_rpsp}. 

\footnotesize
\begin{longtable}{| p{0.15\linewidth} | p{0.1\linewidth} | p{0.45\linewidth} | p{0.25\linewidth} |}
        \toprule
        \hline
        \textbf{Classifiers} & \textbf{Model Families} & \textbf{Description} & \textbf{Language \& Function} \\
        \midrule
        \hline
        \multicolumn{4}{l}{\textbf{1. Discrete Choice Models (3 Models)}} \\
        \hline
        mnl\_B & DCM & Multinomial logit model & Python Biogeme \\
        nl\_B & DCM & Nested logit model & Python Biogeme \\
        mxl\_B & DCM & Mixed logit model (ASC's as random variables) & Python Biogeme \\
        \hline
        \multicolumn{4}{l}{\textbf{2. Deep Neural Networks (15 Models)}} \\
        \hline
        mlp\_R & DNN & Multi-layer perceptrons (MLP) & R RSNNS mlp \\
        avNNet\_R & DNN & Neural network with random seeds with averaged scores; \cite{Ripley1996} & R Caret avNNet \\
        nnet\_R & DNN & Single layer neural network with BFGS algorithm & R Caret nnet \\
        pcaNNet\_R & DNN & PCA pretraining before applying neural networks & R Caret pcaNNet \\
        monmlp\_R & DNN & MLP with monotone constraints \cite{ZhangHong1999} & R Caret monmlp \\
        mlp\_W & DNN & MLP with sigmoid hidden neurons and unthresholded linear output neurons  & Weka MultilayerPerceptron \\
        DNN\_1\_30\_P & DNN & MLP with one hidden layer and 30 neurons in each layer & Python Tensorflow \\
        DNN\_3\_30\_P & DNN & MLP with three hidden layers and 30 neurons in each layer & Python Tensorflow \\
        DNN\_5\_30\_P & DNN & MLP with five hidden layer and 30 neurons in each layer & Python Tensorflow \\
        DNN\_1\_100\_P & DNN & MLP with one hidden layer and 100 neurons in each layer & Python Tensorflow \\
        DNN\_3\_100\_P & DNN & MLP with three hidden layers and 100 neurons in each layer & Python Tensorflow \\
        DNN\_5\_100\_P & DNN & MLP with five hidden layers and 100 neurons in each layer & Python Tensorflow \\
        DNN\_1\_200\_P & DNN & MLP with one hidden layer and 200 neurons in each layer & Python Tensorflow \\
        DNN\_3\_200\_P & DNN & MLP with three hidden layers and 200 neurons in each layer & Python Tensorflow \\
        DNN\_5\_200\_P & DNN & MLP with five hidden layers and 200 neurons in each layer & Python Tensorflow \\
        \hline
        \multicolumn{4}{l}{\textbf{3. Discriminant Analysis (12 Models)}} \\
        \hline        
        lda\_R & DA & Linear discriminant analysis (LDA) model & R Caret lda \\
        lda2\_R & DA & LDA tuning the number of components to retain up to \#classes - 1 & R Caret lda2 \\
        lda\_P & DA & LDA solved by singular value decomposition without shrinkage & Python sklearn LinearDiscriminantAnalysis \\
        sda\_R & DA & LDA with Correlation-Adjusted T (CAT) scores for variable selection  & R Caret sda \\
        lda\_shrink\_P & DA & LDA solved by least squares with automatic shrinkage based on Ledoit-Wolf lemma used. & Python sklearn LinearDiscriminantAnalysis \\
        slda\_R & DA & LDA developed based on left-spherically distributed linear scores & R Caret ipred \\
        stepLDA\_R & DA & LDA model with forward/backward stepwise feature selection  & R Caret stepLDA \\
        pda\_R & DA & Penalized discriminant analysis (PDA) with shrinkage penalty coefficients \cite{Hastie1995} & R mda gen.ridge \\
        mda\_R  & DA & Mixture discriminant analysis (MDA) where the number subclass is tuned to 3 \cite{Hastie1996} & R mda \\
        rda\_R & DA & Regularized discriminant analysis (RDA) with regularized group covariance matrices \cite{Friedman1989} & R klaR \\
        hdda\_R & DA & High dimensional discriminant analysis (hdda)  assuming each class in a Gaussian subspace \cite{Bouveyron2007} & R HD \\
        qda\_P & DA & Quadratic discriminant analysis (qda) & Python sklearn QuadraticDiscriminantAnalysis \\
        \hline
        \multicolumn{4}{l}{\textbf{4. Bayesian Models (7 Models)}} \\
        \hline
        naive\_bayes\_R & BM & Naive Bayes (NB) classifier with the normal kernel density (Laplace correction factor = 2 and Bandwidth Adjustment = 1)  & R naivebayes \\
        nb\_R  & BM & NB classifier with the normal kernel density (Laplace correction factor = 2 and Bandwidth Adjustment = 1)  & R Caret nb \\
        BernoulliNB\_P & BM & NB model  with Bernoulli kernel density function & Python sklearn BermoulliNB \\
        GaussianNB\_P & BM & NB model with Gaussian kernel density function (smoothing = 5, according to the variance portions) & Python sklearn GaussianNB \\
        MultinomialNB\_P & BM & NB model with multinomially distributed data (smoothing = 1 and learn class prior probabilities) & Python sklearn MultinomialNB \\
        
        BayesNet\_W & BM & Bayes network models by hill climbing algorithm \cite{Cooper1992} & Weka BayesNet \\
        NaiveBayes\_W & BM & NB model with Gaussian kernel density function  & Weka NaiveBayes  \\
        \hline
        \multicolumn{4}{l}{\textbf{5. Support Vector Machines (9 Models)}} \\
        \hline
        svmLinear\_R & SVM & Support Vector Machine (SVM) model with linear kernel (inverse kernel width = 1)  & R Caret kernlab \\        
        svmRadial\_R & SVM & Support Vector Machine (SVM) model with Gaussian kernel (inverse kernel width = 1)  & R Caret kernlab \\
        svmPoly\_R  & SVM & SVM with polynomial kernel  & R Caret kernlab \\
        lssvmRadial\_R & SVM & Least Squares SVM model with Gaussian kernel  & R Caret kernlab \\
        LinearSVC\_P & SVM & SVM with linear kernel and l2 penalty  & Python sklearn LinearSVC \\
        SVC\_linear\_P & SVM & SVM with linear kernel (regularization parameter = 1)  & Python sklearn SVC \\
        SVC\_poly\_P & SVM & SVM with polynomial kernel (regularization parameter = 1)  & Python sklearn SVC \\
        SVC\_rbf\_P & SVM & SVM with radial basis function (rbf) kernel (regularization parameter = 1)  & Python sklearn SVC \\   
        SVC\_sig\_P & SVM & SVM with sigmoid function kernel (regularization parameter = 1)  & Python sklearn SVC \\     

        \hline
        \multicolumn{4}{l}{\textbf{6. K Nearest Neighbors (4 Models)}} \\
        \hline
        KNN\_1\_P & KNN & k-nearest neighbors (KNN) classifier with number of neighbors equal to 1  & Python sklearn KNeighborsClassifier \\
        KNN\_5\_P & KNN & KNN classifier with number of neighbors equal to 5  & Python sklearn KNeighborsClassifier \\
        lBk\_1\_W & KNN & KNN classifier with number of neighbors equal to 1 (brute force searching and Euclidean distance) \cite{Aha1991} & Weka lBk \\
        lBk\_5\_W & KNN & KNN classifier with number of neighbors equal to 5 (brute force searching and Euclidean distance) \cite{Aha1991} & Weka lBk \\
        \hline
        \multicolumn{4}{l}{\textbf{7. Decision Tree (14 Models)}} \\
        \hline        
        rpart\_R & DT & Recursive partitioning and regression trees (RPART) model (max depth = 30) & R rpart \\
        rpart2\_R & DT & RPART (max depth = 10)  & R Caret klaR \\
        C5.0Tree\_R & DT & C5.0 decision tree (confidence factor = 0.25) & R Caret C5.0Tree \\
        C5.0Rules\_R & DT & Rule-based models using Quinlan’s C5.0 algorithm \cite{quinlan1993c4} & R Caret C5.0Rules \\
        ctree\_R & DT & Conditional inference trees \cite{Hothorn2006} & R Caret ctree \\
        ctree2\_R & DT & Conditional inference trees (max depth = 10) & R Caret ctree2 \\
        DecisionTree\_P & DT & Decision tree classification model with Gini impurity split measure & Python sklearn DecisionTreeClassifier \\
        ExtraTree\_P & DT & Tree classifier with best splits and features chosen from random splits and randomly selected features \cite{Geurts2006} & Python sklearn ExtraTreeClassifier \\
        DecisionStump\_W & DT & Tree model with decision stump & Weka DecisionStump \\
        HoeffdingTree\_W & DT & An incremental tree with inductive algorithm. \cite{Hulten2001} & Weka HoeffdingTree \\
        REPTree\_W & DT & Tree model using information gain/variance & Weka REPTree \\
        J48\_W & DT & Pruned C4.5 decision tree model & Weka J48 \\
        Attribute Selected\_W & DT & Use J48 trees to classify patterns reduced by attribute selection (Hall, 1998) & Weka AttributeSelected \\
        DecisionTable\_W & DT & Simple decision table majority classier that uses BestFirst as search method \cite{Kohavi1995} & Weka DecisionTable \\
        \hline
        \multicolumn{4}{l}{\textbf{8. Generalized Linear Models (10 Models)}} \\
        \hline        
        Logistic Regression\_l1\_P & GLM & Logistic regression model with l1 penalty & Python sklearn LogisticRegression \\
        Logistic Regression\_l2\_P & GLM & Logistic regression model with l2 penalty & Python sklearn LogisticRegression \\
        Logistic\_W & GLM & Logistic regression model with a ridge estimator \cite{LeCessie1992} & Weka Logistic \\
        SimpleLogistic\_W & GLM & Linear logistic regression models fitted by using LogitBoost \cite{Landwehr2005} & Weka SimpleLogistic \\
        Ridge\_P & GLM & Classifier using Ridge regression & Python sklearn RidgeClassifier \\
        Passive Aggressive\_P & GLM & Passive-aggressive algorithms for classification with hinge loss \cite{Crammer2006} & Python sklearn PassiveAggressiveClassifier  \\
        SGD\_Hinge\_P & GLM & Linear classifier with hinge loss and SGD training & Python sklearn SGDClassifier \\
        SGD\_Squared Hinge\_P & GLM & Linear classifiers of SGD training with squared hinge loss function & Python sklearn SGDClassifier \\
        SGD\_Log\_P & GLM & Linear classifiers of SGD training with log loss function & Python sklearn SGDClassifier \\
        SGD\_Modified Huber\_P & GLM & Linear classifiers of SGD training with modified huber loss function & Python sklearn SGDClassifier \\
        \hline

        \multicolumn{4}{l}{\textbf{9. Gaussian Process (3 Models)}} \\
        \hline        
        GP\_Constant\_P & GP & Gaussian Processes classification model with constant kernel & Python sklearn GaussianProcessClassifier \\
        GP\_DotProduct\_P & GP & Gaussian Processes classification model with Dot-Product kernel & Python sklearn GaussianProcessClassifier \\
        GP\_Matern\_P & GP & Gaussian Processes classification model with Matern kernel & Python sklearn GaussianProcessClassifier \\

        \hline
        \multicolumn{4}{l}{\textbf{10. Bagging (3 Models)}} \\
        \hline
        Bagging\_SVM\_P & BAGGING & A bagging classifier that fits base classifiers based on random subsets of the original dataset; SVM is the base classifier & Python sklearn BaggingClassifier \\
        Bagging\_Tree\_P & BAGGING & A bagging classifier with DecisionTree as the base classifier & Python sklearn BaggingClassifier \\
        Voting\_P & BAGGING & A classifier which combine machine learning classifiers and use a majority vote. We use lda\_P, LinearSVM and Logistic classifiers here. & Python sklearn VotingClassifier \\        
        \hline
        
        \multicolumn{4}{l}{\textbf{11. Random Forests (2 Models)}} \\
        \hline
        RandomForest\_P & RF & A random forest model with 10  trees in the forest & Python sklearn RandomForestClassifier \\
        ExtraTrees\_P & RF & A meta estimator that fits 10 ExtraTree classifiers & Python sklearn ExtraTreeClassifier \\        
        \hline
        \multicolumn{4}{l}{\textbf{12. Boosting (23 Models)}} \\
        \hline
        AdaBoost\_P & BOOSTING & AdaBoost classifier. The DecisionTree with maximum depth =10 is set as the base estimator. \cite{Freund1997} & Python sklearn AdaBoostClassifier \\
        AdaBoostM1\_W & BOOSTING & Boosting method with DecisionStump as the base classifier & Weka AdaboostM1 \\
        AdaBoostM1\_R & BOOSTING & Boosting method with DecisionTree as the base classifier & R adabag Adaboost.M1 \\ 
        LogitBoost\_R & BOOSTING & Logitboost classification algorithm using decision stumps (one node decision trees) as base learners. & R LogitBoost \\        
        
        Gradient Boosting\_P & BOOSTING & An additive model trained in a forward stage-wise fashion \cite{Friedman2001} & Python sklearn GradientBoostingClassifier \\
        DNN\_1\_30\ AdaBoost\_P  & BOOSTING & AdaBoosting method with DNN\_1\_30\_P as the base classifier & Python Tensorflow \\
        DNN\_3\_30\ AdaBoost\_P  & BOOSTING & AdaBoosting method with DNN\_3\_30\_P as the base classifier & Python Tensorflow \\
        DNN\_5\_30\ AdaBoost\_P  & BOOSTING & AdaBoosting method with DNN\_5\_30\_P as the base classifier & Python Tensorflow \\
        DNN\_1\_100\ AdaBoost\_P  & BOOSTING & AdaBoosting method with DNN\_1\_100\_P as the base classifier & Python Tensorflow \\
        DNN\_3\_100\ AdaBoost\_P  & BOOSTING & AdaBoosting method with DNN\_3\_100\_P as the base classifier & Python Tensorflow \\
        DNN\_5\_100\ AdaBoost\_P  & BOOSTING & AdaBoosting method with DNN\_5\_100\_P as the base classifier & Python Tensorflow \\
        DNN\_1\_200\ AdaBoost\_P  & BOOSTING & AdaBoosting method with DNN\_1\_200\_P as the base classifier & Python Tensorflow \\
        DNN\_3\_200\ AdaBoost\_P  & BOOSTING & AdaBoosting method with DNN\_3\_200\_P as the base classifier & Python Tensorflow \\
        DNN\_5\_200\ AdaBoost\_P  & BOOSTING & AdaBoosting method with DNN\_5\_200\_P as the base classifier & Python Tensorflow  \\
        DNN\_1\_30\ GradientBoost\_P  & BOOSTING & Gradient boosting method with DNN\_1\_30\_P as the base classifier & Python Tensorflow \\
        DNN\_3\_30\ GradientBoost\_P  & BOOSTING & Gradient boosting method with DNN\_3\_30\_P as the base classifier & Python Tensorflow \\
        DNN\_5\_30\ GradientBoost\_P  & BOOSTING & Gradient boosting method with DNN\_5\_30\_P as the base classifier & Python Tensorflow \\
        DNN\_1\_100\ GradientBoost\_P  & BOOSTING & Gradient boosting method with DNN\_1\_100\_P as the base classifier & Python Tensorflow \\
        DNN\_3\_100\ GradientBoost\_P  & BOOSTING & Gradient boosting method with DNN\_3\_100\_P as the base classifier & Python Tensorflow \\
        DNN\_5\_100\ GradientBoost\_P  & BOOSTING & Gradient boosting method with DNN\_5\_100\_P as the base classifier & Python Tensorflow \\
        DNN\_1\_200\ GradientBoost\_P  & BOOSTING & Gradient boosting method with DNN\_1\_200\_P as the base classifier & Python Tensorflow \\
        DNN\_3\_200\ GradientBoost\_P  & BOOSTING & Gradient boosting method with DNN\_3\_200\_P as the base classifier & Python Tensorflow \\
        DNN\_5\_200\ GradientBoost\_P  & BOOSTING & Gradient boosting method with DNN\_5\_200\_P as the base classifier & Python Tensorflow  \\
        
        \hline
        \bottomrule
\caption{List of 105 ML classifiers from 12 model families}
\label{table:list_classifiers}
\end{longtable}
\normalsize

This list of models has substantially enriched the models from the literature data. Particularly, the DCM family incorporates MNL, NL, and MXL, thus generating insights from three major DCMs and comparing the performance within the DCM family. Theoretically, we can further enrich the DCMs beyond Table \ref{table:list_classifiers}. For example, the NL models can be expanded using different nest structures; the MXL models by incorporating a flexible correlation matrix or modeling the panel structure in data. It is also possible to expand the substantial scope by examining the departures from purely rational models in choice modelling. However, given the already tremendous scale of our experiments, we decided to use only baseline DCM configurations for simplicity. Nonetheless, the three DCMs are repeatedly trained for various sample sizes, choice categories, and data sets, leading to thousands of experiments and providing sufficiently broad insights and a certain guarantee of robustness in relation to the DCMs' performance. 

\subsection{Contextual factors}\label{set_hyper_dataset}
\noindent
The contextual variables are data sets, sample sizes, and choice categories, which can also influence the model performance besides the model choice. The data sets include NHTS2017, LTDS2015, and SGP2017. NHTS2017 refers to the national household travel survey 2017, collected by the Federal Highway Administration\footnote{Available at https://nhts.ornl.gov/}, which provides travel, vehicle, and individual information across the United States. LTDS2015 refers to the London travel demand survey 2015, which was collected in London with the history of trips from April 2012 to March 2015 and augmented with mode-specific level-of-service variables (e.g. in-vehicle travel time, public transport fares, fuel cost, etc.) \cite{hillel2018recreating}. SGP2017 was collected in 2017 in Singapore through a stated preference survey for travel mode choice \cite{WangShenhao2019_risk_av, shen2019built, mo2020Impacts}, in which the experiments followed the orthogonal survey design \cite{louviere2000stated}. SGP2017 contains information about travel mode choices, socio-demographics, and alternative specific variables. The three data sets are chosen to cover a variety of data collection procedures, geographical locations, and local contexts. The NHTS2017 was collected through revealed preference surveys by the US government, LTDS2015 by combining a transit agency's survey with simulation-based travel information, and SGP2017 by a standard stated preference survey. The three datasets cover the continents of America, Europe, and Asia, thus jointly creating a geographically diverse set. 

The sample sizes span 1,000, 10,000, and 100,000 observations, which are of the same magnitude as the sample sizes from the literature data. When a data set has a large sample size ($>$100,000), it is re-sampled repeatedly to test the impacts of sample size on performance. For example, the NHTS2017 data contains 781,831 trips, among which 1,000, 10,000, and 100,000 observations are repeatedly resampled for predicting travel modes and trip purposes. It also contains 110,565 households, among which the observations are resampled for car ownership prediction. Similar sampling schemes are applied to the LTDS2015 and SGP2017 data sets.

\begin{figure}[H]
\centering
\subfloat[NHTS2017-MC]{\includegraphics[width=0.33\linewidth]{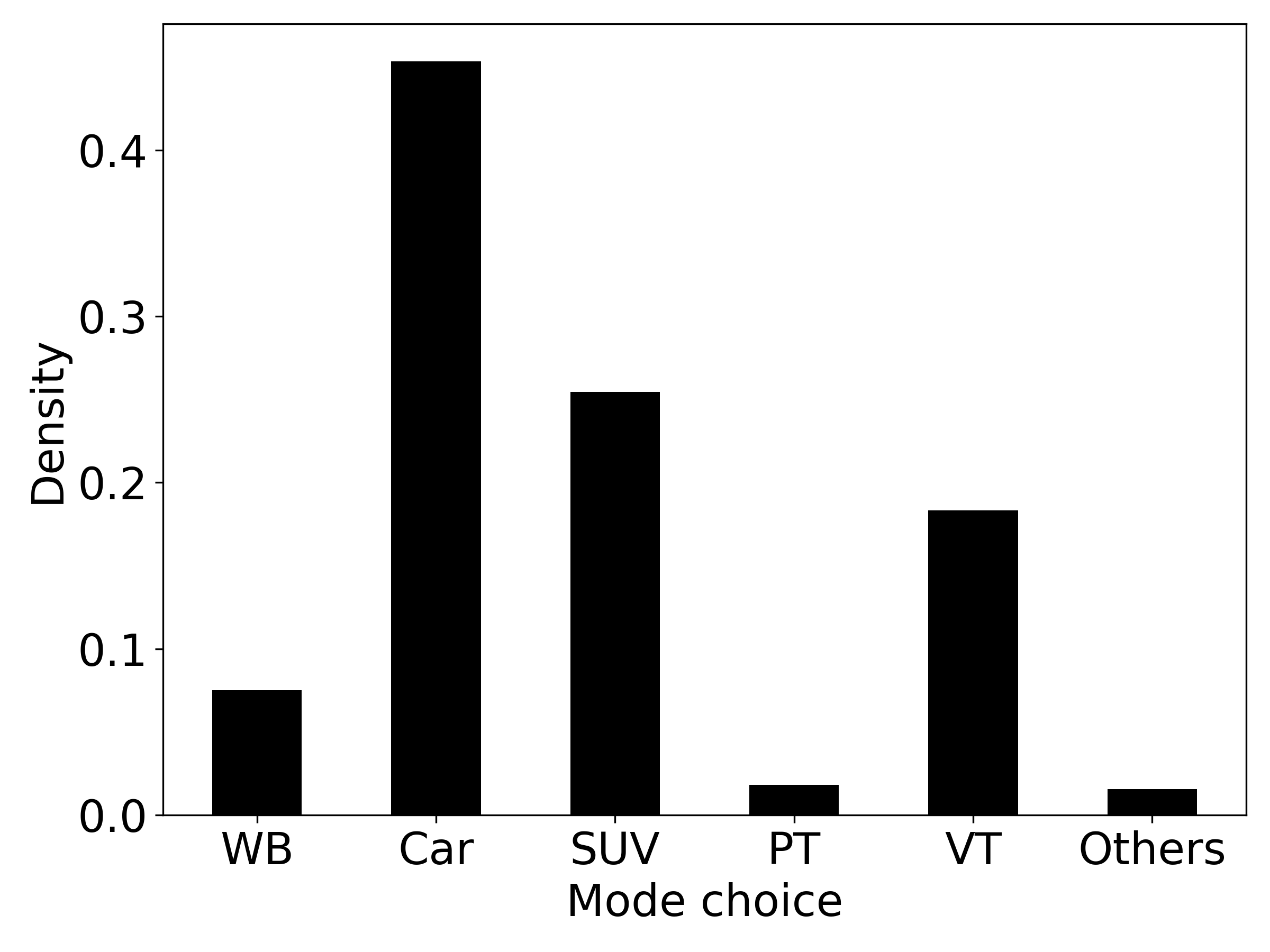}\label{fig_nhts_mc}}
\subfloat[NHTS2017-TP]{\includegraphics[width=0.33\linewidth]{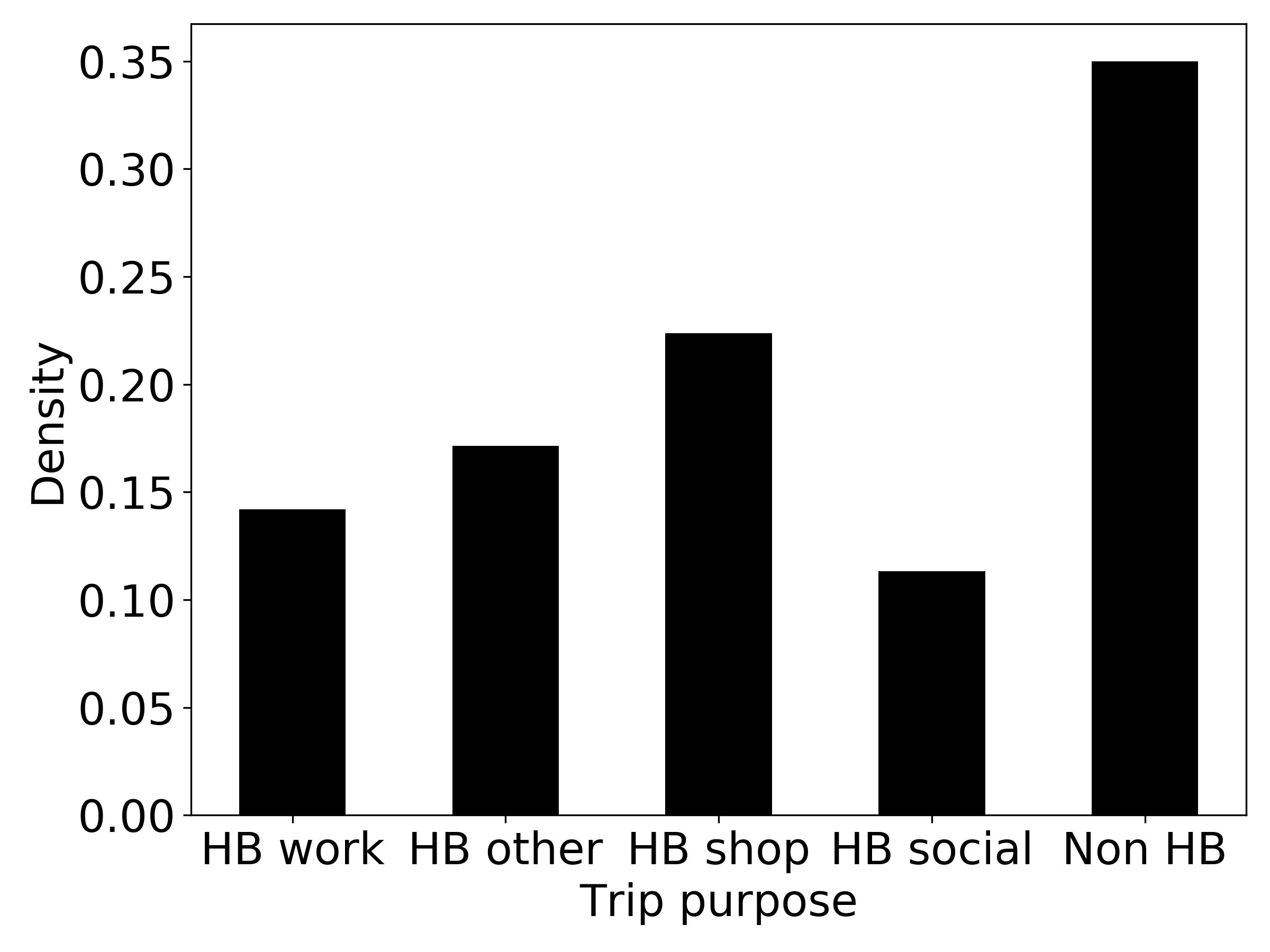}}
\subfloat[NHTS2017-CO]{\includegraphics[width=0.33\linewidth]{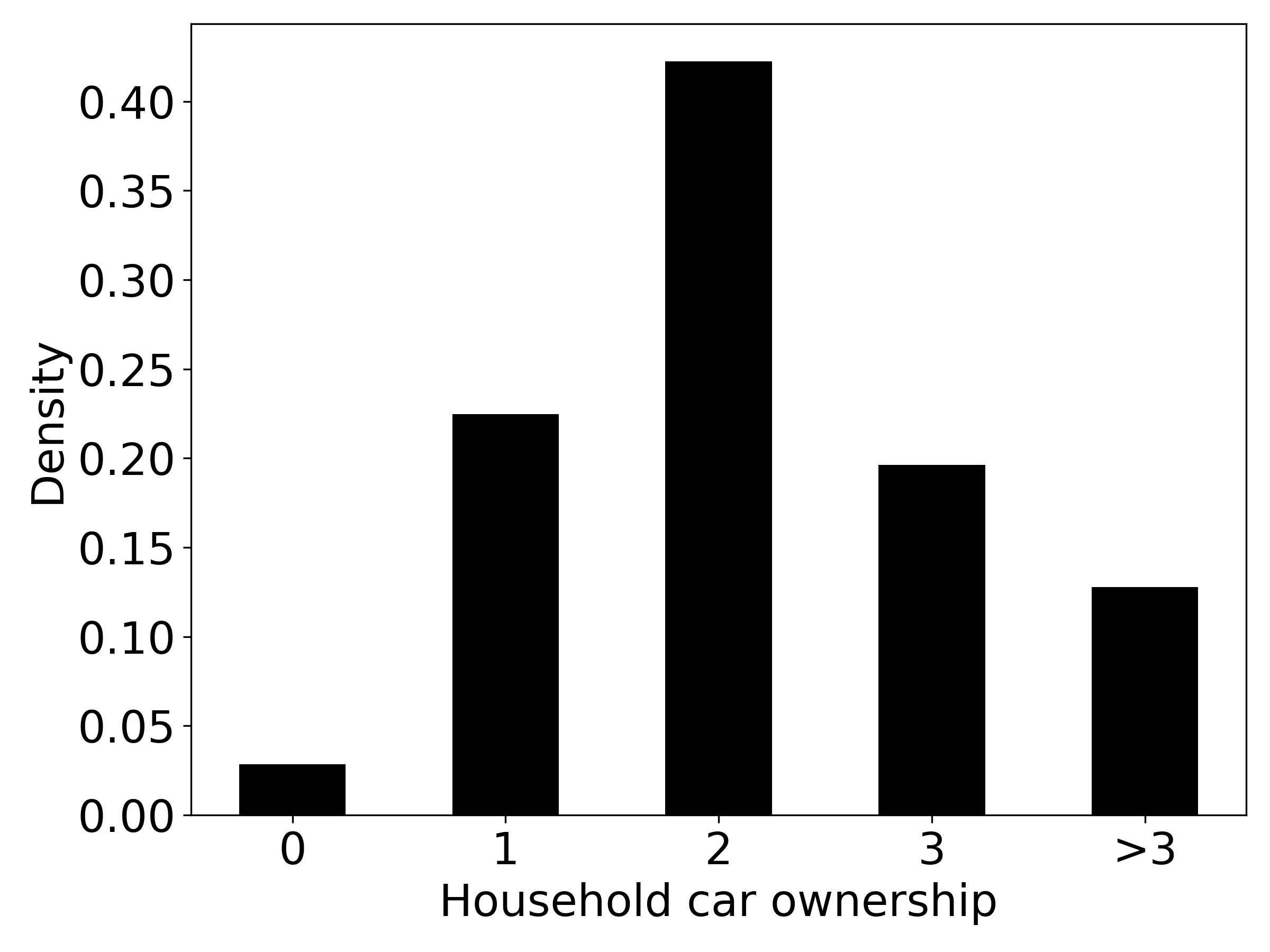}} \\
\subfloat[LTDS2015-MC]{\includegraphics[width=0.33\linewidth]{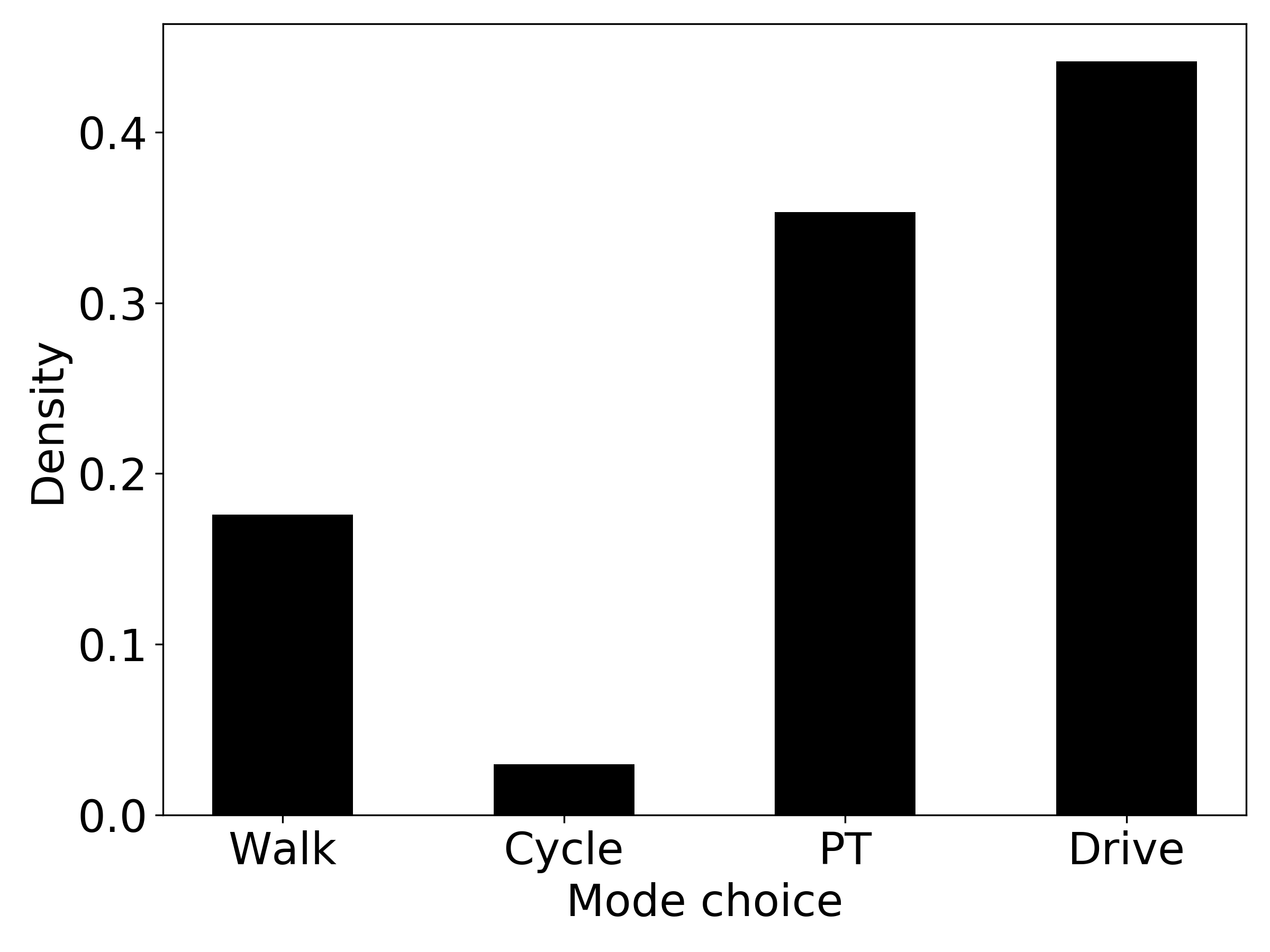}}
\subfloat[SGP2017-MC]{\includegraphics[width=0.33\linewidth]{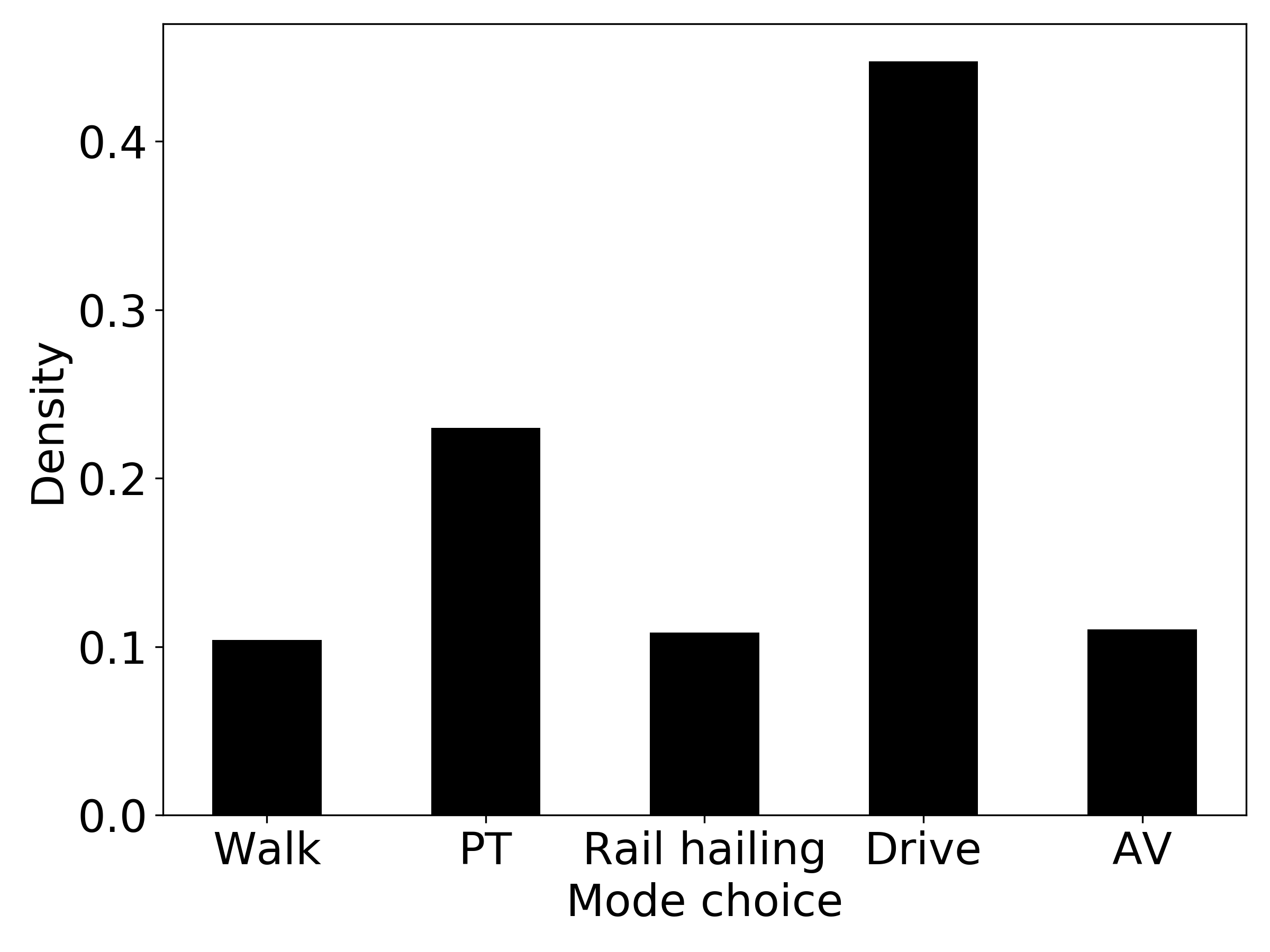}\label{fig_sgp_mc}}
\caption{Output distributions in three datasets}
\label{fig:output_distribution}
\end{figure}

The choice categories include three travel behavior variables: travel modes, trip purposes, and car ownership, which are amongst the most common choice categories from the literature. Travel mode choice is available for all three data sets, while trip purposes and car ownership are only available for the NHTS2017 data. In the NHTS2017 data, the initial 21 travel mode categories are aggregated into six major modes: walk and bike, car, SUV, van and track, public transit, and others. The NHTS2017 data has five trip purposes, including home-based work, home-based shopping, home-based social, home-based others, and none-home-based trips. The car ownership in the NHTS2017 data also has five categories, ranging from zero to more than three cars. The outputs from the three data sets are summarized in Figure \ref{fig:output_distribution}. In the NHTS2017 data, the most common travel mode is car, accounting for 45.3\%, while only 1.8\% of the trips are taken by public transit. In the LTDS2015 and SGP2017 data sets, the largest mode share is driving, which accounts for 44.1\% and 44.7\%, respectively. Public transit is the second largest mode share in London and Singapore, accounting for 35.3\% and 23.0\% of the total trips. 

Due to the large scale of our experiments, we have to somewhat compromise the completeness of our experiments for successful implementation. We chose the most important input features from three data sets using $\chi^2$ feature selection, without further randomization in feature selection algorithms. Although the theoretical maximal number of experiments equals to $105 \times 3 \times 3 \times 3 \times 5 = 14,175$, we completed only $6,970$ experiments due to limitations in data availability, storage, and computational difficulty (See Appendix II for details). 

\section{Results}
\label{sec:results}
\subsection{Tournament models}
\noindent Tables \ref{table:tournament_literature} and \ref{table:tournament_experiment} present the results of tournament models for the literature and the experiment data. In both tables, Models 1-3 correspond to the regressions using only model-family effects, combining model-family and contextual factors, and adding interactions of model-family and contextual factors. The main dimension of interests is the twelve model families represented by eleven dummy variables with DCM as the baseline category. Hence the $\beta's$ measure the difference of the other families from the DCMs. The contextual variables include data sources, choice categories, input and output dimensions, and sample size. In both tables, Model 2 is used mainly for model interpretation because the linear model-family and contextual effects are directly interpretable, while Model 3 is designed to showcase the potential improvement in the fitting of the tournament model. For computational efficiency, we randomly sampled 2,000 pairs of models from the two data sources as inputs of the tournament models for fair comparison. The tournament models yield four findings below. 

\begin{table}[th!]
\centering 
\caption{Results of Tournament Models (literature data)}
\resizebox{0.95\linewidth}{!}{

\begin{tabular}{llll}
\toprule
\multirow{2}{*}{\textbf{Parameters}} & \multicolumn{3}{c}{\textbf{Tournament Models}}                                                                                                                 \\ \cmidrule(l){2-4} 
                                     & \multicolumn{1}{l}{1. Model families} & \multicolumn{1}{l}{2. With contexts} & \multicolumn{1}{l}{3. With interactions} \\ \midrule
\multicolumn{4}{l}{\textit{Model family effects ($\beta$)}}                                                                                                                                          \\
DCM                              & Fix to 0                   & Fix to 0                  & Fix to 0                                        \\
                                 BAGGING & 2.870 (0.746) *** &  4.630 (0.840) *** &      1.110 (1.330) \\
                                       BM & -1.030 (0.317) ** &     -0.183 (0.375) & -3.800 (1.040) *** \\
                                 BOOSTING & 1.050 (0.138) *** &  1.910 (0.187) *** &   -2.040 (0.970) * \\
                                      DNN &  0.270 (0.096) ** &  0.921 (0.124) *** &   -2.150 (0.975) * \\
                                       DT & 1.280 (0.126) *** &  1.450 (0.166) *** &   -1.920 (0.965) * \\
                                      GLM &    -7.380 (12.30) &     -5.730 (18.80) &     -8.790 (16.70) \\
                                      KNN &     0.096 (0.544) &      0.973 (0.631) &     -2.230 (1.160) \\
                                       RF & 1.520 (0.162) *** &  2.300 (0.217) *** &     -1.370 (0.989) \\
                                      SVM & 1.010 (0.125) *** &  1.110 (0.155) *** &   -2.170 (0.998) * \\
\multicolumn{4}{l}{\textit{Contextual effects ($\gamma$)}}\\
                                 If\_US & / & Fix to 0 &     Fix to 0 \\
                              If\_Asian &                 / & -0.559 (0.122) *** &     -0.233 (0.156) \\
                                   If\_AM &                 / & -2.200 (0.353) *** & -2.850 (0.386) *** \\
                               If\_Europe &                 / & -1.790 (0.142) *** & -1.830 (0.196) *** \\
                                 If\_TM & / & Fix to 0 &     Fix to 0 \\
                                   If\_CO &                 / & -2.840 (0.250) *** & -2.510 (0.295) *** \\
                                   If\_TP &                 / & -1.520 (0.183) *** & -1.610 (0.228) *** \\
                          Input dimension &                 / & -0.067 (0.008) *** &                  / \\
                      Num of alternatives &                 / & -0.442 (0.038) *** &                  / \\
              Sample size ($\times 10^5$) &                 / &     -0.067 (0.097) &                  / \\
                    Input dimension (DCM) &                 / &                  / &      0.097 (0.059) \\
              (Input dimension (DCM))$^2$ &                 / &                  / &     -0.003 (0.002) \\
                     Input dimension (ML) &                 / &                  / & -0.272 (0.034) *** \\
               (Input dimension (ML))$^2$ &                 / &                  / &  0.008 (0.001) *** \\
                Num of alternatives (DCM) &                 / &                  / & -4.040 (0.373) *** \\
          (Num of alternatives (DCM))$^2$ &                 / &                  / &  0.355 (0.036) *** \\
                 Num of alternatives (ML) &                 / &                  / & -1.390 (0.233) *** \\
           (Num of alternatives (ML))$^2$ &                 / &                  / &  0.104 (0.022) *** \\
        Sample size ($\times 10^5$) (DCM) &                 / &                  / &      0.098 (0.881) \\
  (Sample size ($\times 10^5$) (DCM))$^2$ &                 / &                  / &     -0.052 (0.437) \\
         Sample size ($\times 10^5$) (ML) &                 / &                  / &  -1.300 (0.448) ** \\
   (Sample size ($\times 10^5$) (ML))$^2$ &                 / &                  / &   0.673 (0.214) ** \\
\midrule
\multicolumn{4}{l}{\textit{Statistical summary}}\\
                       Num of Observations &              2000 &               2000 &               2000 \\
                     Final log likelihood &        -1,233.868 &           -916.866 &           -802.047 \\
                      Null log likelihood &        -1,386.294 &         -1,386.294 &         -1,386.294 \\
                                      AIC &         2,487.736 &          1,869.731 &          1,658.094 \\
                                 $\rho^2$ &              0.11 &              0.339 &              0.421 \\
                        Adjusted $\rho^2$ &             0.103 &              0.326 &              0.402 \\
Average probability of correct prediction &              0.57 &                0.7 &              0.739 \\
                                 Hit rate &             0.652 &              0.774 &              0.792 \\
\bottomrule
\end{tabular}
}
\label{table:tournament_literature}
\end{table}

First, we find that many ML models indeed outperform DCMs, as suggested by the statistical significance from the positive coefficients of the ML model families. In the literature data (Columns 1 and 2 in Table \ref{table:tournament_literature}), the majority of the ML families have positive coefficients with high levels of statistical significance. Specifically, bagging, boosting, DNNs, decision trees, and random forests can outperform DCMs in predicting travel behavior in a statistically significant manner with and without controlling the contextual variables. By comparing the magnitudes of the coefficients (e.g., Model 2), we can rank the performance from the highest to the lowest as bagging, random forest, boosting, decision tree, DNNs, DCMs, etc. This finding is further reinforced in our experiment data. In Columns 1-2 of Table \ref{table:tournament_experiment}, the ensemble methods and DNNs continue to outperform the benchmark DCMs, although the statistical significance is much lower than the literature data. Since the model-family coefficients only reveal the average effects across model families, we will delve into the variance in and the best models of the model families in the sections below. In both literature and experiment data, naive bayesian models and Gaussian process underperform DCMs. The performance of K-nearest neighbors and support vector machines is inconclusive due to their inconsistent signs of coefficients across the literature and experiment data. In short, our tournament models demonstrate that many ML models, particularly ensemble methods and DNNs, can statistically outperform DCMs in predicting travel behavior, consistent with the majority of the previous deterministic comparisons. 

However, our results also demonstrate that the contextual effects ($\gamma's$) are more dominating than the choice of models ($\beta's$) in influencing model performance. In Model 1 of Tables \ref{table:tournament_literature} and \ref{table:tournament_experiment}, where contextual controls are not imposed, the pseudo $R^2$ reaches only 0.11 and 0.06. But after adding the contextual variables (Model 2 in Tables \ref{table:tournament_literature} and \ref{table:tournament_experiment}), the pseudo $R^2$ increases drastically to 0.34 and 0.43. This explanatory power of the contextual variables is also demonstrated by their statistical significance, since the contextual variables are much more significant than the model variables. Nearly all the contextual variables are statistically significant in literature and experiment data, while only around 50\% of the model family coefficients are statistically significant. The coefficients of the contextual effects suggest that the model performance in the US context is typically higher than Asian and European contexts, and that the predictive performance in travel mode choice is higher than that in trip purposes and car ownership, at least in the literature data. The contextual effects also suggest that the intrinsic predictive power decreases with a higher number of alternatives and input dimensions, and increases with larger sample size. A large number of alternatives lead to challenges in model prediction, and large sample size facilitates the parameter training, thus improving model performance. Different from our expectations, high input dimensions reduce model performance, so we investigate such an issue in Appendix VI. Meanwhile, the contextual variables and the model families appear to exhibit some interaction effect, because the contextual variables can amplify the model family effects. Model 3 in Tables \ref{table:tournament_literature} and \ref{table:tournament_experiment} demonstrate that such interaction effects exist, since Model 3 further improves the pseudo $R^2$ to 0.42 and 0.45. This finding highlights the significant contextual variations in model performance, implying that it might be inappropriate to abstractly compare model performance without explicitly examining the contextual effects. 

\begin{table}[th!]
\centering
\caption{Results of Tournament Models (experiment data)}
\resizebox{0.95\linewidth}{!}{
\begin{tabular}{llll}
\toprule
\multirow{2}{*}{\textbf{Parameters}} & \multicolumn{3}{c}{\textbf{Tournament Models}}                                                                                                                 \\ \cmidrule(l){2-4} 
& \multicolumn{1}{l}{1. Model families} & \multicolumn{1}{l}{2. With contexts} & \multicolumn{1}{l}{3. With interactions} \\ \midrule
\multicolumn{4}{l}{\textit{Model family effects ($\beta$)}}                                                                                                                                          \\
DCM                              & Fix to 0                   & Fix to 0                  & Fix to 0                                        \\
                                 BAGGING &      0.278 (0.285) &     -0.138 (0.385) &    0.552 (0.253) * \\
                                       BM & -0.966 (0.251) *** & -2.010 (0.354) *** & -1.430 (0.207) *** \\
                                 BOOSTING &      0.271 (0.219) &      0.221 (0.301) &  0.826 (0.130) *** \\
                                       DA &     -0.131 (0.223) &   -0.659 (0.309) * &     -0.073 (0.146) \\
                                      DNN &    0.468 (0.226) * &      0.482 (0.312) &  1.100 (0.148) *** \\
                                       DT &      0.184 (0.224) &      0.064 (0.306) &  0.684 (0.135) *** \\
                                      GLM &      0.127 (0.225) &     -0.120 (0.308) &  0.546 (0.147) *** \\
                                       GP &   -0.909 (0.404) * & -1.740 (0.528) *** &   -0.837 (0.403) * \\
                                      KNN &  -0.790 (0.272) ** & -1.620 (0.374) *** & -1.020 (0.240) *** \\
                                       RF &      0.388 (0.306) &      0.371 (0.416) &  1.120 (0.298) *** \\
                                      SVM &  -0.748 (0.245) ** & -1.600 (0.340) *** & -0.981 (0.182) *** \\
\multicolumn{4}{l}{\textit{Contextual effects ($\gamma$)}}\\
If\_US & / & Fix to 0 &     Fix to 0 \\
                                If\_Asian &                  / & -0.599 (0.033) *** & -0.715 (0.050) *** \\
                               If\_Europe &                  / &  1.100 (0.091) *** &  0.899 (0.112) *** \\
If\_TM & / & Fix to 0 &     Fix to 0 \\                               
                                   If\_CO &                  / &  0.553 (0.102) *** &      0.122 (0.103) \\
                                   If\_TP &                  / &   -0.217 (0.099) * & -0.654 (0.106) *** \\
                          Input dimension &                  / &  -0.013 (0.005) ** &                  / \\
                      Num of alternatives &                  / & -1.750 (0.091) *** &                  / \\
              Sample size ($\times 10^5$) &                  / &  0.786 (0.115) *** &                  / \\
                    Input dimension (DCM) &                  / &                  / &     -0.149 (0.289) \\
              (Input dimension (DCM))$^2$ &                  / &                  / &      0.002 (0.004) \\
                     Input dimension (ML) &                  / &                  / &   -0.114 (0.054) * \\
               (Input dimension (ML))$^2$ &                  / &                  / &    0.002 (0.001) * \\
                Num of alternatives (DCM) &                  / &                  / &     -0.284 (1.130) \\
          (Num of alternatives (DCM))$^2$ &                  / &                  / &     -0.173 (0.153) \\
                 Num of alternatives (ML) &                  / &                  / &     -0.834 (1.120) \\
           (Num of alternatives (ML))$^2$ &                  / &                  / &     -0.130 (0.110) \\
        Sample size ($\times 10^5$) (DCM) &                  / &                  / &    16.70 (8.490) * \\
  (Sample size ($\times 10^5$) (DCM))$^2$ &                  / &                  / &     -14.60 (8.030) \\
         Sample size ($\times 10^5$) (ML) &                  / &                  / &  10.30 (1.390) *** \\
   (Sample size ($\times 10^5$) (ML))$^2$ &                  / &                  / & -9.120 (1.320) *** \\
   \midrule
\multicolumn{4}{l}{\textit{Statistical summary}}\\
                      Num of Observations &               2000 &               2000 &               2000 \\
                     Final log likelihood &         -1,293.057 &            -789.98 &           -762.561 \\
                      Null log likelihood &         -1,386.294 &         -1,386.294 &         -1,386.294 \\
                                      AIC &          2,610.114 &          1,617.960 &          1,581.123 \\
                                 $\rho^2$ &              0.067 &               0.43 &               0.45 \\
                        Adjusted $\rho^2$ &              0.059 &              0.416 &               0.43 \\
Average probability of correct prediction &              0.545 &              0.748 &              0.759 \\
                                 Hit rate &              0.617 &              0.832 &               0.84 \\
\bottomrule
\end{tabular}
}
\label{table:tournament_experiment}
\end{table}

Third, even after incorporating all the model and contextual factors, the tournament model exhibits significant residual randomness, implying that inherent uncertainty could exist in such model comparisons. In Model 3, the pseudo $R^2$ of the tournament models reaches only 0.42 and 0.45 for the literature and experiment data. It suggests that at least half of variation in model comparisons remains purely random. This residual randomness could be attributed to the lack of critical contextual variables, incomplete model specification, or inherent randomness in such comparison tasks. Although it is always possible to further improve the explanatory power by incorporating more factors or enriching model specification, our finding suggests at least that researchers should not simply conclude the superiority of ML methods over DCMs without in-depth statistical analysis. Uncertainty prevails in such model comparisons, and researchers could seek to quantify the confidence in the comparative results. 

Lastly, besides the consistent patterns between the literature and experiment results, we also identify many inconsistent patterns, which could reveal potential publication biases. On average, ML models have more positive coefficients and higher statistical significance in the literature data than the experiment data. For example, the coefficients of bagging, boosting, and random forest in literature data are 4.63, 1.91, and 2.30 (Table 3), which are much larger than -0.14, 0.22, and 0.37 in the experiment data (Table 4). Meanwhile, the three coefficients are statistically significant at the 99.99\% confidence level in the literature data, while they are insignificant in the experiment data. This finding implies that the published studies are more likely than our experiments to identify ML models as the dominating ones. Such publication biases could be caused by researchers' self-filtering or journals' preferences for ML models' dominating performance over traditional DCMs. Regarding the contextual variables, both data sets indicate that it becomes harder to compare models in travel purpose predictions and with higher dimensional outputs. But the location and other contextual effects are inconsistent between the two data sources, which implies some difficulty in identifying consistent contextual effects for model performance. 

The tournament model summarizes ranking of model pairs, but it does not provide direct insights into prediction accuracy itself. In Appendix III, we introduce a simple linear model with prediction accuracy as outputs, which yield similar conclusions to the tournament models above. Although the tournament models can efficiently summarize experiments on average, they might conceal the rich contents in the variation of the model performance, particularly because modelers often compare across the best models from each model family, rather than the average ones. For example, various DNN architectures are lumped into the DNN family, and the MNL, NL, and MXL models lumped into a single DCM family. Therefore, we will examine the model performance along individual dimensions using visualization and descriptive statistics, thus further distinguishing the differences within and across model families in the two sections below. 

\subsection{Examining performance across model families and individual models}
\begin{figure}[th!]
\centering
{\includegraphics[width=1.0\linewidth]{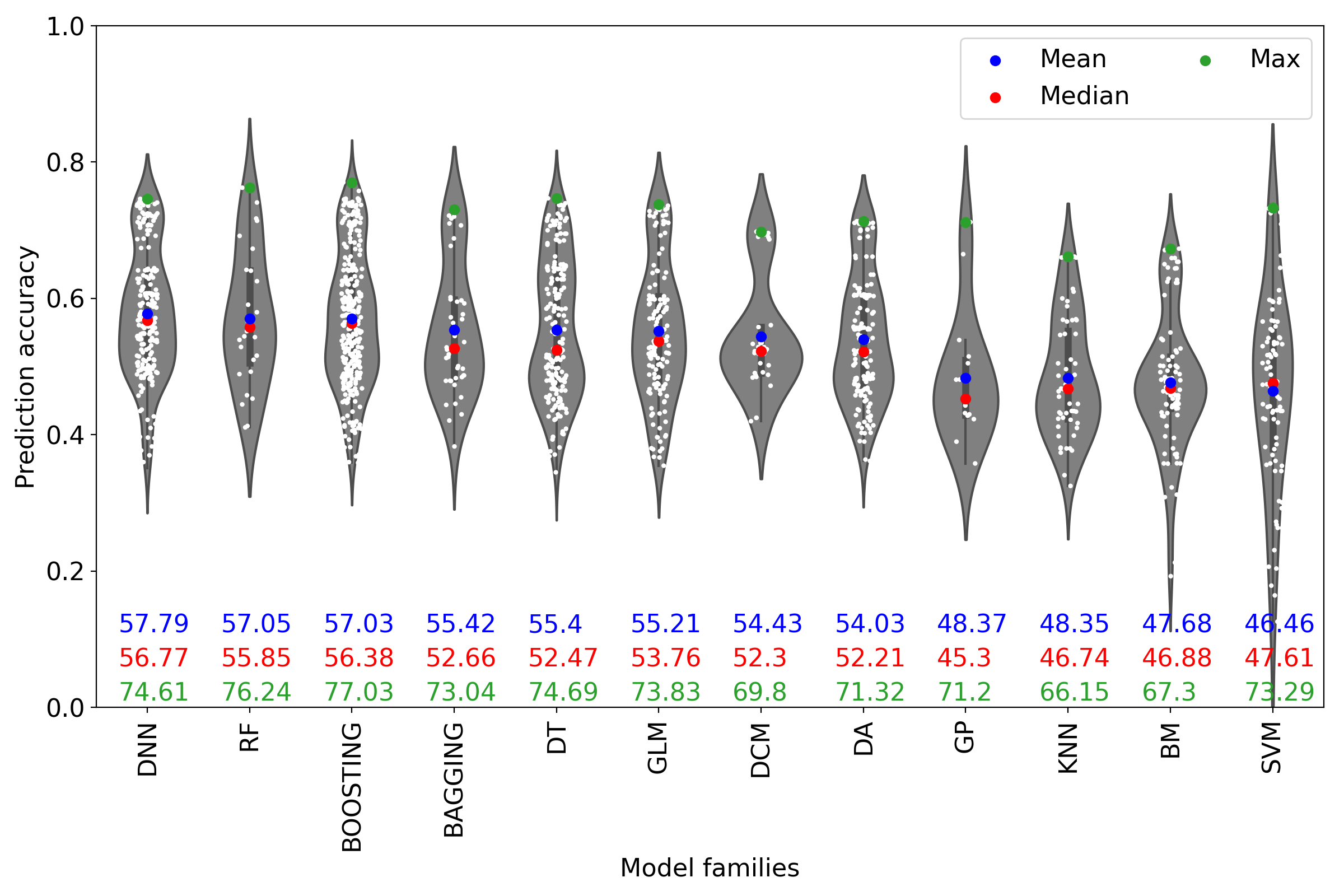}}\\
\caption{Prediction accuracy of 12 model families}
\label{fig:accuracy_model_families}
\end{figure}

\begin{figure}[th!]
\centering
{\includegraphics[width=1.0\linewidth]{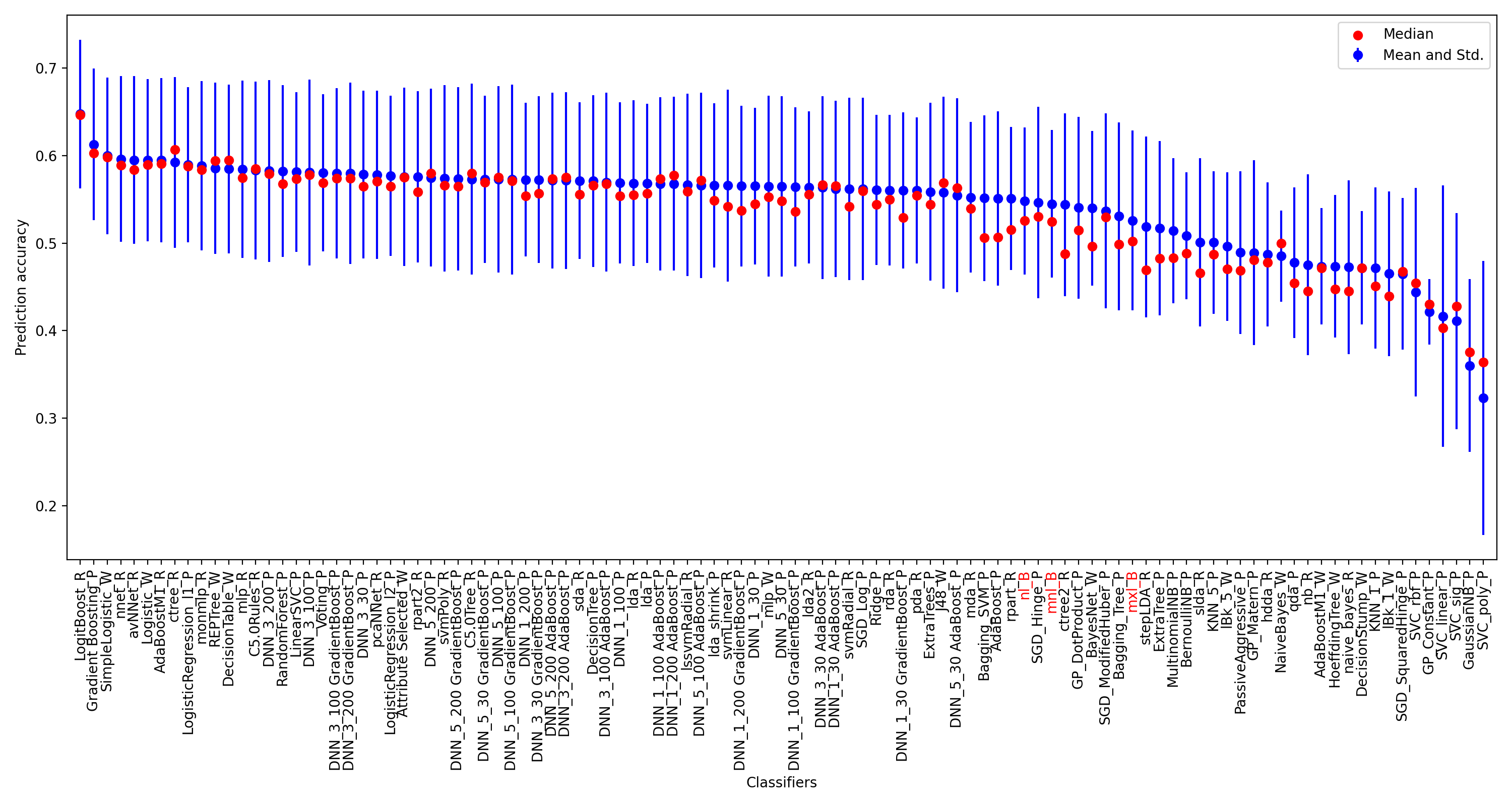}}\\
\caption{Prediction accuracy of 105 models}
\label{fig:accuracy_models}
\end{figure}

\noindent Figures \ref{fig:accuracy_model_families} and \ref{fig:accuracy_models} present visualizations of prediction accuracy for 12 model families and 105 individual models. Both figures are sorted based on average prediction accuracy, from highest to lowest. In Figure \ref{fig:accuracy_model_families}, the x-axis represents model families, while the y-axis represents prediction accuracy. Each white dot represents an experiment's accuracy, and the grey areas indicate density distributions of predictive performance. The blue, red, and green dots represent the mean, median, and maximum prediction accuracy, respectively. Figure \ref{fig:accuracy_models} extends the 12 model families to 105 classifiers using a similar format, replacing the violin plots with simple blue bars to depict performance variance. In this figure, the blue and red dots respectively represent the mean and median prediction accuracy. Additionally, Appendix IV contains an analysis of literature data, yielding similar findings to the experiment data presented here.

Figure \ref{fig:accuracy_model_families} illustrates the relative prediction accuracy of different model families. Among them, ensemble methods and DNNs exhibit relatively high prediction accuracy, while the DCM family ranks 7th out of 12. The DNN family, comprising 15 classifiers, achieves mean and median accuracies of 57.79\% and 56.77\% respectively. The three ensemble methods, namely random forests, boosting, and bagging, also demonstrate high accuracy with mean values of 57.05\%, 57.03\%, and 55.42\% respectively. Despite incorporating more complex error structures in the NL and MXL models, the DCMs still fail to outperform the ensemble methods and DNNs. The superior prediction accuracy of ML models can be attributed to their strong approximation power \cite{Hornik1989} and effective regularization techniques \cite{Srivastava_Hinton2014, Boyd2004}. This visualization further confirms the findings from our previous tournament model, providing additional ranking information for each model family.

Figure \ref{fig:accuracy_model_families} also reveals significant variation and even multi-modal distributions in predictive performance. For instance, the DNN family exhibits an average accuracy of approximately 57.79\%, while its maximum prediction accuracy reaches around 74.61\%. Similarly, the average accuracy of the DCM family stands at 54.43\%, which is 15.37\% lower than its maximum value of 69.8\%. Remarkably, each model family can achieve maximum values in the range of 70-80\%, surpassing the average by 20-30 percentage points. Several model families display explicit multi-modal distributions, including DCM and BAGGING. The wide variation and multi-modal distributions in prediction accuracy visually demonstrate that the choice of models may not be the most crucial factor in determining the final predictive performance. This observation aligns with our previous finding regarding the substantial uncertainty inherent in such model comparisons. 

Figure \ref{fig:accuracy_models} demonstrates that individual models do not deviate significantly from the average performance of their respective model families, and substantial variation persists within each of the 105 individual models. Among these models, the top performers are LogitBoosting and GradientBoosting from the BOOSTING family, followed by DNNs such as nnet, avNNet, and monmlp, which rank 4th, 5th, and 10th out of 105 models. Within the DCM model family, MNL, NL, and MXL (highlighted in red) exhibit similar performance at the medium to lower end of the spectrum among the 105 models. Their average prediction accuracies are 54.50\%, 55.01\%, and 52.58\%, respectively, only slightly deviating from the DCM family's average. However, each individual model represents a significant variation, with performance varying by at least 20 percentage points, as indicated by the long blue bars. The fact that MXL performs worst may come as a surprise to readers, given that in estimation, the inclusion of random heterogeneity almost invariably leads to big gains in performance. However, here, we are talking about prediction performance, which implies averaging again over the random components, and this removes much of the differences between MXL and non-mixture DCM models.

\subsection{Examining performance across data sets, outputs, and sample sizes}
\label{sec_hyper_dataset}
\begin{figure}[H]
\centering
{\includegraphics[width=0.75\linewidth]{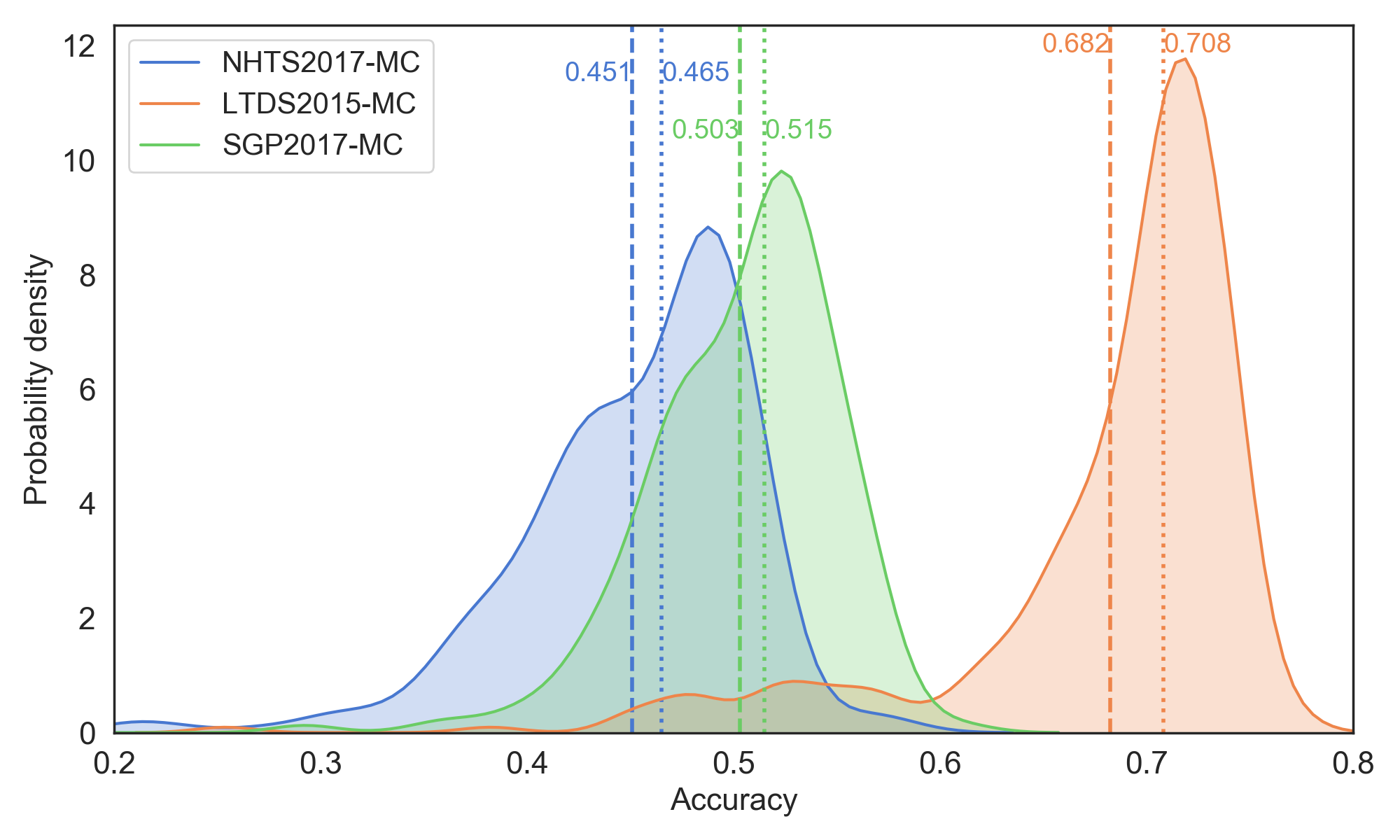}}\\
\caption{Prediction accuracy by data sets (dashed and dotted lines represent mean and median values)}
\label{fig:KDE}
\end{figure}

\begin{figure}[H]
\centering
\subfloat[NHTS2017-MC]{\includegraphics[width=0.2\linewidth]{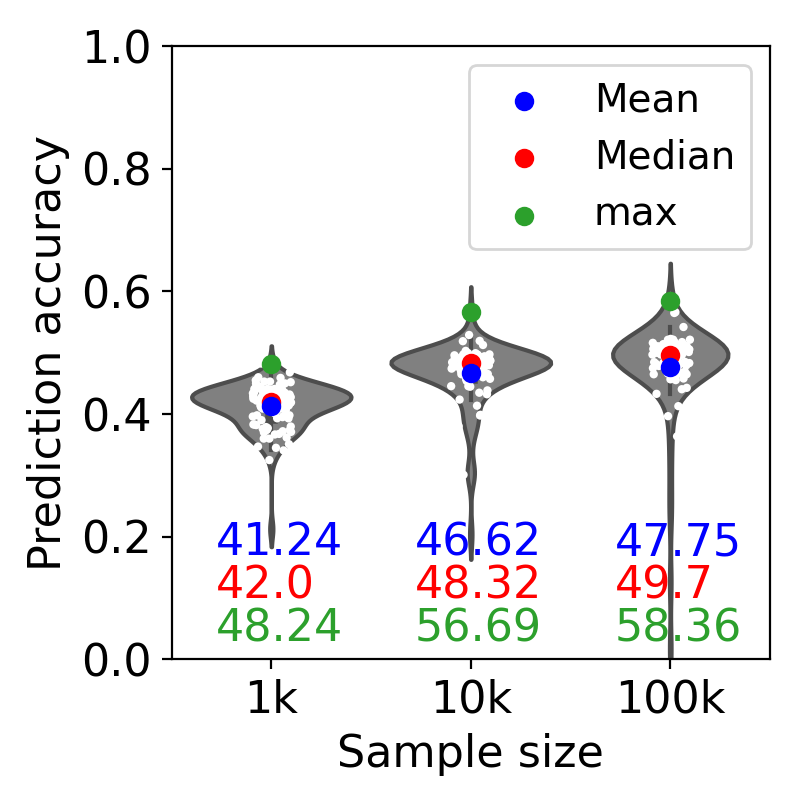}\label{sfig:Sample_size_Accuracy_NHTS_MODE}}
\subfloat[NHTS2017-TP]{\includegraphics[width=0.2\linewidth]{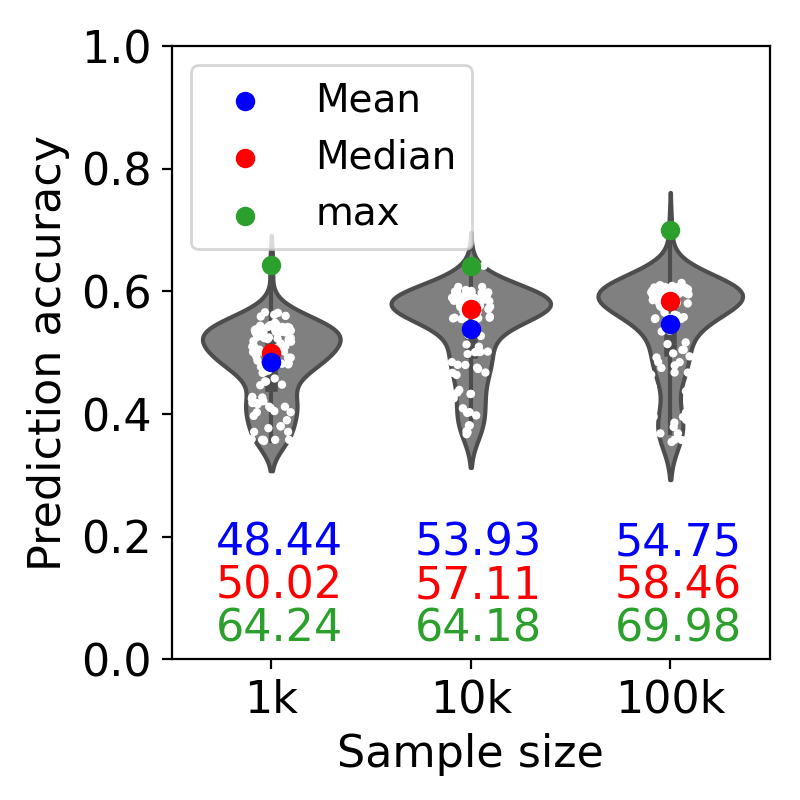}\label{sfig:Sample_size_Accuracy_NHTS_TP}}
\subfloat[NHTS2017-CO]{\includegraphics[width=0.2\linewidth]{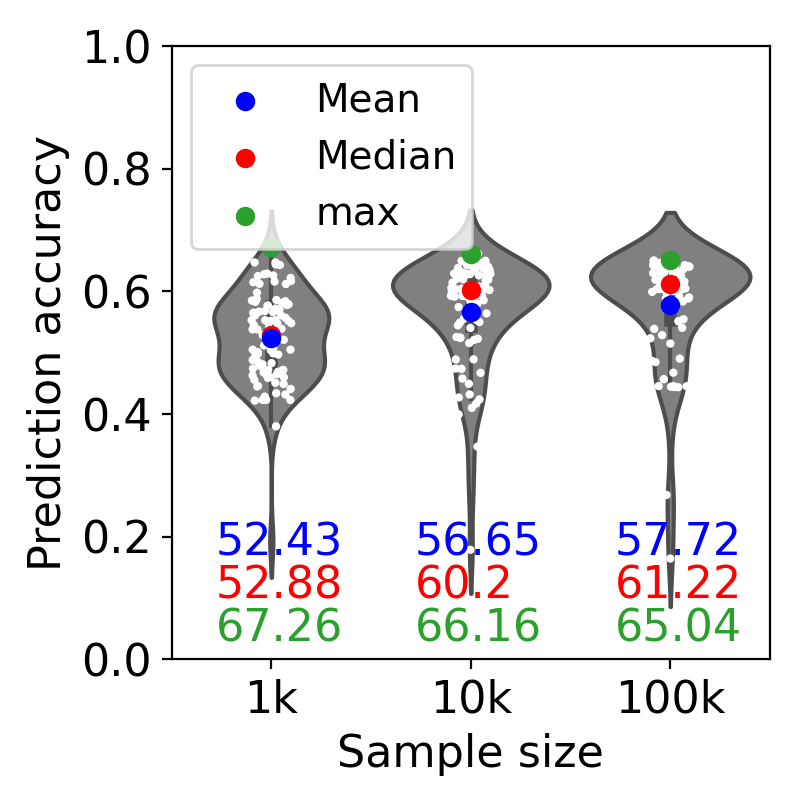}\label{sfig:Sample_size_Accuracy_NHTS_CO}}
\subfloat[LTDS2015-MC]{\includegraphics[width=0.2\linewidth]{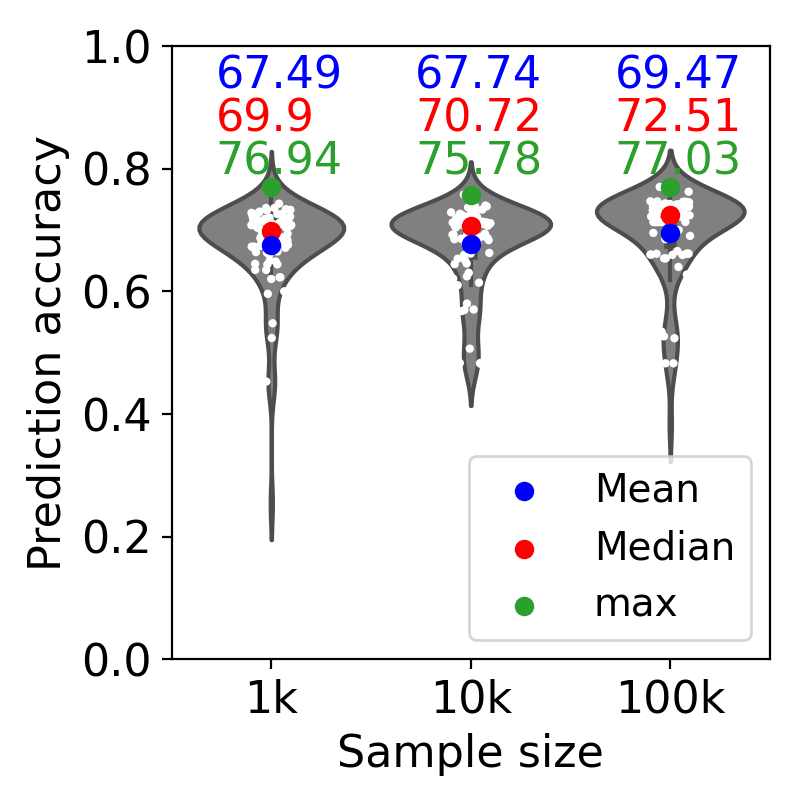}}
\subfloat[SGP2017-MC]{\includegraphics[width=0.2\linewidth]{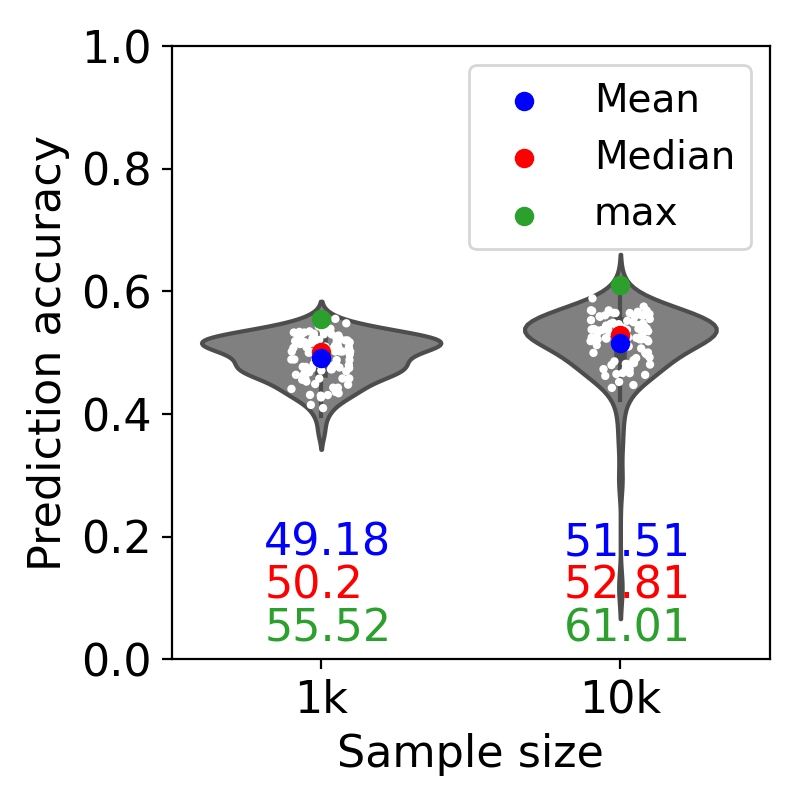}\label{sfig:Sample_size_Accuracy_SG_MODE}}
\caption{Prediction accuracy by sample size, datasets, and outputs}
\label{fig:accuracy_sample_size_by_tasks}
\end{figure}

\noindent
The substantial uncertainty within model families and individual models demonstrates the importance of investigating the contextual factors' impacts on model performance. Therefore, we further decompose the predictive performance along the contextual dimensions. Figure \ref{fig:KDE} visualizes the distributions of prediction accuracy in three data sets, and Figure \ref{fig:accuracy_sample_size_by_tasks} examines the prediction accuracy regarding data sets, outputs, and sample sizes. Table \ref{table:list_topN_dataset} reports the model ranking and performance in three panels and nine sub-tables. Within each sub-table, we report the top-10 models with the DCMs highlighted below, conditioning on specific data sets, outputs, and sample sizes. With such analysis, we could examine the performance consistency of individual models and model families across contexts.

Figures \ref{fig:KDE} and \ref{fig:accuracy_sample_size_by_tasks} demonstrate that predictive performance exhibits significant variation across datasets and shows a moderate increase with larger sample sizes. While model families display multi-modal performance distributions in Figure \ref{fig:accuracy_model_families}, the performance distributions of datasets appear more concentrated around the average in Figure \ref{fig:KDE}. The highest accuracy is attained in LTDS2015 (76.58\%), which is approximately 20 percentage points higher than NHTS2017 and SGP2017 data. The mean and median values of all three distributions are close, indicating a relatively concentrated distribution within each data set. Furthermore, predictive performance generally improves with sample size, although the rate of improvement appears to diminish slightly. For example, when the sample size is 1,000, the mean prediction accuracy is 51.63\%, which increases to around 55.29\% and 57.42\% as the sample size grows to 10,000 and 100,000. This trend of marginally diminishing improvement holds consistent across datasets and outputs, as demonstrated in all five subfigures of Figure \ref{fig:accuracy_sample_size_by_tasks}.

\begin{table}[th!]
\centering
\caption{Model performance specific to data sets, outputs, and sample sizes}
\resizebox{1.0\linewidth}{!}{
\begin{tabular}{ccc|ccc|ccc}
\hline
\multicolumn{9}{l}{\textbf{Panel 1. Performance across data sets}} \\
\hline
\multicolumn{3}{c|}{NHTS2017-MC}            & \multicolumn{3}{c|}{LTDS2015-MC}            & \multicolumn{3}{c}{SGP2017-MC}            \\ \hline
Rank & Model                     & Accuracy & Rank & Model                     & Accuracy & Rank & Model                   & Accuracy \\ \hline
1    & LogitBoost\_R             & 0.5443   & 1    & LogitBoost\_R             & 0.7658   & 1    & LogitBoost\_R           & 0.5827   \\
2    & Gradient Boosting\_P      & 0.5130   & 2    & Gradient Boosting\_P      & 0.7431   & 2    & avNNet\_R               & 0.5564   \\
3    & SimpleLogistic\_W         & 0.5001   & 3    & avNNet\_R                 & 0.7386   & 3    & nnet\_R                 & 0.5552   \\
4    & LogisticRegression\_l1\_P & 0.4919   & 4    & nnet\_R                   & 0.7351   & 4    & Gradient Boosting\_P    & 0.5550   \\
5    & Logistic\_W               & 0.4904   & 5    & SimpleLogistic\_W         & 0.7350   & 5    & DNN\_1\_200\_P          & 0.5487   \\
6    & ctree\_R                  & 0.4881   & 6    & Logistic\_W               & 0.7339   & 6    & DNN\_1\_100 AdaBoost\_P & 0.5484   \\
7    & DecisionTable\_W          & 0.4861   & 7    & svmPoly\_R                & 0.7329   & 7    & DNN\_3\_100 AdaBoost\_P & 0.5481   \\
8    & monmlp\_R                 & 0.4834   & 8    & lssvmRadial\_R            & 0.7320   & 8    & DNN\_3\_100\_P          & 0.5480   \\
9    & LinearSVC\_P              & 0.4822   & 9    & monmlp\_R                 & 0.7317   & 9    & DNN\_1\_100\_P          & 0.5474   \\
10   & REPTree\_W                & 0.4808   & 10   & LogisticRegression\_l1\_P & 0.7312   & 10   & DNN\_5\_200 AdaBoost\_P & 0.5474   \\ \hline
36   & nl\_B                     & 0.4674   & 71   & mxl\_B                    & 0.6980   & 34   & nl\_B                   & 0.5270   \\
41   & mnl\_B                    & 0.4657   & 76   & nl\_B                     & 0.6933   & 35   & mnl\_B                  & 0.5262   \\
94   & mxl\_B                    & 0.4250   & 77   & mnl\_B                    & 0.6917   & 44   & mxl\_B                  & 0.5200   \\ 
\hline
\hline
\multicolumn{9}{l}{\textbf{Panel 2. Performance across outputs}} \\
\hline
\multicolumn{3}{c|}{MC (NHTS2017+LTDS2015+SGP2017)} & \multicolumn{3}{c|}{NHTS2017-TP}            & \multicolumn{3}{c}{NHTS2017-CO}        \\ \hline
Rank    & Model                       & Accuracy   & Rank & Model                     & Accuracy & Rank & Model                 & Accuracy \\ \hline
1       & LogitBoost\_R               & 0.6369     & 1    & LogitBoost\_R             & 0.6613   & 1    & LogitBoost\_R         & 0.6615   \\
2       & Gradient Boosting\_P        & 0.6098     & 2    & Gradient Boosting\_P      & 0.5901   & 2    & rpart2\_R             & 0.6464   \\
3       & SimpleLogistic\_W           & 0.5953     & 3    & LogisticRegression\_l1\_P & 0.5859   & 3    & AdaBoostM1\_R         & 0.6464   \\
4       & avNNet\_R                   & 0.5926     & 4    & ctree\_R                  & 0.5825   & 4    & Attribute Selected\_W & 0.6459   \\
5       & nnet\_R                     & 0.5921     & 5    & mlp\_R                    & 0.5809   & 5    & DecisionTable\_W      & 0.6450   \\
6       & Logistic\_W                 & 0.5920     & 6    & SimpleLogistic\_W         & 0.5802   & 6    & Gradient Boosting\_P  & 0.6440   \\
7       & LogisticRegression\_l1\_P   & 0.5894     & 7    & avNNet\_R                 & 0.5774   & 7    & ctree\_R              & 0.6400   \\
8       & monmlp\_R                   & 0.5882     & 8    & Logistic\_W               & 0.5771   & 8    & REPTree\_W            & 0.6380   \\
9       & AdaBoostM1\_R               & 0.5834     & 9    & LogisticRegression\_l2\_P & 0.5768   & 9    & SimpleLogistic\_W     & 0.6315   \\
10      & LinearSVC\_P                & 0.5832     & 10   & Voting\_P                 & 0.5762   & 10   & nnet\_R               & 0.6301   \\ \hline
52      & nl\_B                       & 0.5670     & 52   & nl\_B                     & 0.5473   & 81   & mnl\_B                & 0.5124   \\
57      & mnl\_B                      & 0.5656     & 69   & mnl\_B                    & 0.5228   & 82   & nl\_B                 & 0.5078   \\
73      & mxl\_B                      & 0.5477     & 74   & MXL\_B                    & 0.5020   & 86   & mxl\_B                & 0.4840   \\ 
\hline
\hline
\multicolumn{9}{l}{\textbf{Panel 3. Performance across sample sizes}} \\
\hline
\multicolumn{3}{c|}{1k}                      & \multicolumn{3}{c|}{10k}               & \multicolumn{3}{c}{100k}              \\ \hline
Rank & Model                     & Accuracy & Rank & Model                & Accuracy & Rank & Model                & Accuracy \\ \hline
1    & LogitBoost\_R             & 0.6244   & 1    & LogitBoost\_R        & 0.6476   & 1    & LogitBoost\_R        & 0.6760   \\
2    & Gradient Boosting\_P      & 0.5874   & 2    & Gradient Boosting\_P & 0.6205   & 2    & Gradient Boosting\_P & 0.6353   \\
3    & AdaBoostM1\_R             & 0.5772   & 3    & avNNet\_R            & 0.6082   & 3    & C5.0Rules\_R         & 0.6303   \\
4    & SimpleLogistic\_W         & 0.5752   & 4    & monmlp\_R            & 0.6070   & 4    & ctree\_R             & 0.6295   \\
5    & svmPoly\_R                & 0.5741   & 5    & nnet\_R              & 0.6068   & 5    & DNN\_5\_200\_P       & 0.6261   \\
6    & nnet\_R                   & 0.5724   & 6    & Logistic\_W          & 0.6054   & 6    & monmlp\_R            & 0.6257   \\
7    & avNNet\_R                 & 0.5708   & 7    & SimpleLogistic\_W    & 0.6049   & 7    & DNN\_3\_200\_P       & 0.6253   \\
8    & lssvmRadial\_R            & 0.5666   & 8    & DNN\_3\_200\_P       & 0.6001   & 8    & Logistic\_W          & 0.6248   \\
9    & LogisticRegression\_l1\_P & 0.5664   & 9    & DNN\_3\_100\_P       & 0.5974   & 9    & REPTree\_W           & 0.6244   \\
10   & svmLinear\_R              & 0.5658   & 10   & ctree\_R             & 0.5972   & 10   & DNN\_5\_100\_P       & 0.6244   \\ \hline
50   & mnl\_B                    & 0.5270   & 14   & DNN\_5\_200\_P       & 0.5943   & 70   & nl\_B                & 0.5683   \\
52   & nl\_B                     & 0.5256   & 64   & nl\_B                & 0.5600   & 73   & mnl\_B               & 0.5584   \\
57   & DNN\_5\_200\_P            & 0.5144   & 67   & mnl\_B               & 0.5523   & -    & -                    & -        \\
\hline
\end{tabular}
}
\label{table:list_topN_dataset}
\end{table}

Table \ref{table:list_topN_dataset} demonstrates that the relative ranking of individual models remains relatively stable across different contexts, while their absolute prediction accuracy could vary. When considering specific data sets (Panel 1), ensemble methods and DNNs still achieve the highest performance. LogitBoost and GradientBoosting demonstrate dominant performance across all three data sets, and DNNs like avNNet, nnet, and DNN\_200\_1 often rank among the top 5 models in LTDS2015 and SGP2017 datasets. This observation holds true in Panels 2 and 3, where LogitBoost and GradientBoosting achieve the highest predictive performance across sample sizes and outputs. DNNs and DTs frequently emerge as top-10 models across all three panels. Meanwhile, the DCMs (MNL, MXL, and NL models) consistently perform at the medium to lower end of the performance distributions in all nine contexts. When conditioning on sample sizes or outputs, the MNL, NL, and MXL models fail to achieve high-quality performance, ranking around the 50-70th position for each sample size.

Apart from the general patterns, the authors find two intriguing observations. Firstly, as the sample size increases, the relative ranking of DCMs and DNNs exhibits an opposite trend: DNNs' ranking improves while DCMs' ranking declines. In Panel 3, when the sample size grows from 1,000 to 100,000, the relative ranking of MNL and NL drops from 50th to 73rd. Conversely, the DNN\_200\_5 model achieves a prediction accuracy of approximately 51.4\% at a sample size of 1,000, but its ranking jumps significantly from 57th to 5th when the sample size reaches 100K. Secondly, the relative ranking of MNL, NL, and MXL is not consistent across contexts. In the NHTS2017-MC data, MXL performs worse than MNL and NL, while it outperforms them in the LTDS2015-MC data. Occasionally, the simplest model (e.g., MNL) attains the highest performance (Panel 2, NHTS2017-CO). This discrepancy may arise because performance evaluation occurs in the testing set, where more complex models do not necessarily outperform simpler ones. Despite the specifics, these two observations support the main finding regarding significant contextual randomness in model performance. Although rankings already offer more stability than absolute prediction accuracy, the ranking of individual models still varies depending on contextual factors such as sample sizes and data sources.

\section{Limitations}
\label{sec:limitations}
\noindent 
This benchmark study has important limitations regarding data sources, experiment design, evaluation metrics, and research scope. While our literature and experiment data are relatively comprehensive, they are not exhaustive. The literature data excludes previous studies focusing solely on ML or DCM and those centered on spatiotemporal travel data \cite{Mozolin2000, Polson2017, WuYuankai2018}. Our experiments exclude certain dimensions, such as training algorithms, hyper-parameters, feature selection, and feature transformation, which could influence model performance and change the benchmark results in our tournament model. Additionally, this empirical benchmark study uses prediction accuracy and model ranking as the evaluation criteria, neglecting other metrics like log-likelihood or Brier score \cite{Abdar2021, Gruber2022}. Probabilistic metrics could be more appropriate for model evaluation from a theoretical perspective. However, since log-likelihood as a probabilistic metric does not exist in many deterministic ML models, prediction accuracy is the only metric that enables our comprehensive comparison of both deterministic and probabilistic models. The authors compared only DNNs and DCMs using log-likelihood and Brier scores in Appendix V, with a more comprehensive evaluation with probabilistic metrics left for future studies. Furthermore, model comparison should not solely focus on predictive performance. Recent studies have explored innovations in deep learning architectures \cite{Sifringer2020, Han2022, Wong2021}, investigated the potential of images and natural language \cite{Cranenburgh2022, Wang2023}, and examined interpretability and robustness as critical dimensions for model evaluation \cite{Lipton2016, WangShenhao2020_econ_info, Wang2021_dnn_stat_learning}. Lastly, although we find the potential publication biases in favor of ML models' higher performance by comparing literature and experiment data, our own experiment data are not devoid of publication biases. Publication bias appears an inherently challenging topic because any manuscript, including the unpublished ones, contains the authors' intention of being published, and the existing tests for publication biases are only statistical heuristics, which are largely inapplicable to the ML models \cite{Thornton2000}. Unfortunately, these limitations cannot be addressed in this empirical benchmark work and need be investigated by future researchers.

\section{Conclusions}
\label{sec:conclusions}
\noindent
Recent studies comparing DCMs and ML models have often overlooked the randomness in model performance and have been limited to specific contexts, featuring a small number of models and data sets. To overcome these challenges, this study presents an empirical benchmark that introduces a tournament model, examining both literature and experiment data. This approach allows for formal statistical modeling to capture the inherent randomness in model comparisons and summarize results from a large number of experiments. Our data sources are extensive, comprising 6,970 experiments from 105 models and 12 model families in our experiment data, and summarizing 136 experiments from 35 previous studies in the literature data. Our benchmark study yields two findings. 

On the one hand, our benchmark study confirms that ML models generally outperform DCMs. The tournament models reveal positive and statistically significant coefficients for ML model families, indicating the superiority of ML models in predictive performance. Ensemble methods such as boosting, bagging, and random forests, along with DNNs, consistently demonstrate the highest predictive performance. In contrast, the DCM family consistently falls within the medium to lower range of the predictive distribution. This trend holds across both literature and experiment data, with or without contextual factors as controls. This consistent pattern applies to both aggregate model families and individual models, as individual model performance aligns closely with the average of their respective model families. The ML models, such as LogitBoost and GradientBoosting, consistently achieve dominating performance across all data sets, sample sizes, and choice categories. Overall, our findings reinforce and further validate previous conclusions using formal statistical analysis.

However on the other hand, our benchmark study emphasizes the influence of contextual factors and the residual randomness inherent in model comparisons. It becomes evident that contextual factors play a more significant role in explaining the outcomes of model comparisons than the choice of models themselves. Statistical significance is observed for all contextual factors in our models, whereas the significance of model families is comparatively weaker. Even after considering all model and contextual variables, a considerable amount of residual randomness persists. This residual randomness may arise from incomplete input dimensions or inherent complexity in model comparisons. Furthermore, our descriptive analysis also reveals significant randomness, as demonstrated by the wide variation in predictive performance among model families and individual models. These findings suggest that the outcomes of performance comparisons are primarily shaped by contextual factors and intrinsic randomness, highlighting the limited perspective of exclusively focusing on model choice. 

This empirical benchmark study improves upon the research frontier of the deterministic model comparisons through innovations in methodology, findings, and implications. Regarding methodology, the tournament model emulates and yet generalizes the past deterministic comparisons. It resembles the comparative studies because modelers also seek to identify the model of higher performance through pairwise comparison. Yet, this tournament model generalizes the deterministic comparison by expanding the sample size, examining the impacts of contextual complexity, and adopting formal statistical methods to investigate the effects of model families for both literature and experiment data points. From a modeler's perspective, such a probabilistic approach captures the multiple sources of randomness in model comparison, including data collection, selection of training and testing data, stochastic algorithms, and inherent randomness in every predictive task \cite{Dietterich1998}. In fact, statistical meta-modeling approaches have been widely adopted in the machine learning community for comparing classifiers across data sets, with methods including pairwise comparison \cite{Bradley1952_rank_analysis}, block-wise tests \cite{Demsar2006_stat_comparison}, or recently Bayesian approaches \cite{Wainer2023_bayesian_comparison}. From a mathematical perspective, the past deterministic comparisons are only specific cases of our tournament models. The tournament model degenerates into a deterministic comparative study when the variance of $\epsilon_m$ approaches zero, the contextual factors are removed, and the sample size of pairwise comparisons is reduced to only one observation. In fact, if a deterministic process represents the underlying data generating process, our probabilistic tournament model should yield very high predictive power (e.g., 100\% $\rho^2$ when no omitted variable exists). However, our empirical finding suggests otherwise, reaching a $\rho^2$ of only 0.4. Such an empirical result suggests that the performance comparison is much more random than a pure deterministic process. 

Regarding findings, the tournament model reveals the generalizable patterns between the literature and the experiment data. By compiling two comprehensive data sets, the authors identify the shared trends across a wide range of experiments, surpassing the limitations of past context-specific analyses. ML and DCMs exhibit relatively consistent rankings across different contexts, and the overall impacts of contextual factors and intrinsic randomness outweigh the influence of model choice in determining model performance. Such findings are corroborated in both data sets.
Regarding implications, this work can be continuously updated and widely adopted to evaluate the performance of a large number of models. Researchers could use the tournament models to evaluate model performance, thus partially avoiding the strenuous efforts of training models and collecting data. The coefficients for model families and contextual factors can be used as the prior beliefs in a Bayesian or a Frequentist framework. By augmenting new literature or experiment results, future researchers can track the changes of the model-family or contextual effects in the upcoming years. To facilitate future researchers to replicate our work, we uploaded the scripts and data sets to a Github repository\footnote{\url{https://github.com/mbc96325/Comparing-machine-learning-and-discrete-choice-model}}.

Our study highlights two important directions for future research: model transferability and uncertainty quantification. While previous deterministic comparisons have generally yielded correct conclusions, they may overlook the significance of contextual factors when fixed within a specific dataset. To address this, future studies could focus on model transferability across contexts by testing models across diverse datasets or by examining the influence of varying inputs and outputs within the same dataset. Additionally, the presence of significant residual randomness in model comparisons underscores the need to unravel the uncertainty structures in ML models. Unlike statistical methods, current ML models in practice provide deterministic predictions without statistical properties that characterize data or model uncertainty. This deterministic approach often leads to overly simplified conclusion that ML models outperform DCMs without providing statistical confidence in such a statement. Therefore, future research should explore methods for quantifying the uncertainty and testing the transferability for ML models, thus generating robust and generalizable findings regarding the predictive power of ML and DCMs in travel demand analysis.

\section*{Acknowledgement} 
\noindent
We thank Singapore-MIT Alliance for Research and Technology (SMART) for partially funding this research, and thank Alex Guo for his collection of the references in the literature review. Shenhao thanks Stefanie Jegelka to provide key guidance about this paper idea in a casual talk in 2017. Stephane Hess acknowledges support from the European Research Council through the consolidator grant 615596-DECISIONS and advanced grant 101020940-SYNERGY. 

\section*{CRediT Author Statement}
\noindent
The authors confirm contribution to the paper as follows. \textbf{Shenhao Wang:} Conceptualization, Data curation, Formal analysis, Investigation, Methodology, Project administration, Validation, Writing (original draft), and Writing (review and editing). \textbf{Baichuan Mo:} Data curation, Formal analysis, Investigation, Methodology, Software, and Visualization. \textbf{Yunhan Zheng:}  Formal analysis, Investigation, and Software. \textbf{Stephane Hess:} Writing (review and editing). \textbf{Jinhua Zhao:} Funding acquisition and Resources.

\bibliographystyle{vancouver}
\bibliography{main}

\end{document}